\documentclass[11pt]{article}
\usepackage{amsmath}
\usepackage[utf8]{inputenc}
\usepackage{booktabs}
\usepackage[preprint]{acl}

\usepackage{times}
\usepackage{latexsym}
\usepackage{amssymb}
\usepackage{pgfplots}
\usepackage[most]{tcolorbox}
\usepackage{listings}
\usepackage{setspace}
\usepackage{enumitem}
\usepackage{ragged2e}
\usepackage{etoolbox}

\lstdefinestyle{wraptt}{
  basicstyle=\ttfamily\footnotesize,
  breaklines=true,
  breakatwhitespace=false,
  columns=fullflexible,
  keepspaces=true,
  showstringspaces=false,
  tabsize=2,
  frame=single,
  framerule=0.4pt
}

\setlist[itemize]{nosep,left=1.2em}

\newtcolorbox{promptbox}[1]{
  breakable,
  enhanced jigsaw,
  colback=gray!2,
  colframe=gray!55,
  title=\textbf{#1},
  fonttitle=\bfseries,
  boxrule=0.5pt,
  toptitle=2pt,
  bottomtitle=2pt,
  left=6pt,right=6pt,top=6pt,bottom=6pt,
}

\newcommand{\code}[1]{\texttt{#1}}

\pgfplotsset{compat=1.18}

\usepackage[T1]{fontenc}

\usepackage[utf8]{inputenc}

\usepackage{microtype}

\usepackage{inconsolata}

\usepackage{graphicx}

\title{\fontsize{14}{16}\selectfont Collaborative Multi-Agent Test-Time Reinforcement Learning for Reasoning}

\author{
  Zhiyuan Hu$^{1,2}$\thanks{Zhiyuan Hu. \href{mailto:hzycs@mit.edu}{Email: hzycs@mit.edu}} \quad 
  Yunhai Hu$^{3}$ \quad 
  Juncheng Liu$^{4}$ \quad 
  Shuyue Stella Li$^{5}$ \quad 
  Yucheng Wang$^{2}$ \quad 
  \textbf{Zhen Xu}$^{6}$ \quad \\
  \textbf{See-Kiong Ng}$^{2}$ \quad 
  \textbf{Anh Tuan Luu}$^{7}$ \quad
  \textbf{Xinxing Xu}$^{4}$ \quad
  \textbf{Bryan Hooi}$^{2}$ \quad
  \textbf{Cynthia Breazeal}$^{1}$ \quad 
  \textbf{Hae Won Park}$^{1}$ \quad \\
  \textsuperscript{1} MIT \quad
  $^2$ NUS \quad
  $^3$ NYU \quad
  $^4$ Microsoft \quad
  $^5$ UW \quad
  $^6$ Columbia \quad
  $^7$ NTU
}

\usepackage{graphicx}
\usepackage{setspace}
\usepackage{tikz}
\usetikzlibrary{positioning}
\tikzset{
  mybox/.style={draw, fill=yellow!20, thick, rectangle, rounded corners, inner sep=10pt, inner ysep=10pt},
  fancytitle/.style={fill=black!10, font=\bfseries, inner sep=3pt}
}
\begin{document}
\maketitle
\begin{abstract}

Multi-agent systems have evolved into practical LLM-driven collaborators for many applications, gaining robustness from diversity and cross-checking. However, multi-agent RL (MARL) training is resource-intensive and unstable: co-adapting teammates induce non-stationarity, and rewards are often sparse and high-variance. 
Therefore, we introduce \textbf{Multi-Agent Test-Time Reinforcement Learning (MATTRL)}, a framework that injects structured textual experience into multi-agent deliberation at inference time. MATTRL forms a multi-expert team of specialists for multi-turn discussions, retrieves and integrates test-time experiences, and reaches consensus for final decision-making. 
We also study credit assignment for constructing a turn-level experience pool, then reinjecting it into the dialogue.
Across challenging benchmarks in medicine, math, and education, MATTRL improves accuracy by an average of 3.67\% over a multi-agent baseline, and by 8.67\% over comparable single-agent baselines. Ablation studies examine different credit-assignment schemes and provide a detailed comparison of how they affect training outcomes. MATTRL offers a stable, effective and efficient path to distribution-shift-robust multi-agent reasoning without tuning. Code can be found here.\footnote{\url{https://github.com/zhiyuanhubj/MATTRL}}


\end{abstract}

\section{Introduction}


Multi-agent systems have moved from early algorithmic prototypes to practical LLM-driven collaborators.
Across math, coding, web interaction, and analytical benchmarks, these multi-agent systems reliably outperform comparable single-agent baselines, as diversity and cross-checking improve robustness under distribution shift.

Recent works explore collaborative multi-agent frameworks to enhance LLM agents’ capabilities. For example, AutoGen \cite{wu2024autogen} (orchestrated multi-agent dialogues with tool use and human-in-the-loop), CAMEL \cite{li2023camel} (role-playing with inception prompting), AgentVerse \cite{chen2023agentverse} (an open platform for cooperative problem solving and social simulation), ChatDev \cite{qian2023chatdev} (specialized software agents for design, coding, and testing), and Magentic-One \cite{fourney2024magentic} (an orchestrator that routes tasks among specialized agents for web/local workflows). In parallel, the success of DeepSeek-R1 \cite{guo2025deepseek} has catalyzed reinforcement learning (RL) as a post-training paradigm for stronger reasoning. 
Efforts to extend RL to the multi-agent setting include MAPoRL \cite{park2025maporl}, which jointly optimizes multi-model discussions and final answers via RL, and ReMA \cite{wan2025rema}, which separates high-level meta-thinking from low-level reasoning into two agents and trains them with GRPO.

However, MARL remains resource-intensive and can erode general abilities when adapted to a single domain. Training stability is also difficult to guarantee due to (i) non-stationarity from simultaneously evolving teammates, which shifts state and return distributions, and (ii) sparse, high-variance rewards. 
Hence, we propose Multi-Agent Test-Time Reinforcement Learning (MATTRL), an adaptation framework that injects test-time textual experience into the collaborative process. Instead of updating weights, MATTRL conditions behavior with structured experience, enabling rapid, distribution-shift-robust adaptation to new tasks/domains without harming original generality. Additionally, textual experience provides richer turn-level signals about collaboration quality and reasoning than scalar rewards alone.
Textual experience mitigates key MARL pain points by keeping policies fixed and providing dense, stepwise experience at every turn.

The crucial components of MATTRL include (1) various group-to-agent credit assignment strategies for experience selection, (2) construction of an experience pool from test time examples, and (3) integration of the experience pool into the multi-agent collaborative process. 
MATTRL first instantiates a team of specialized agents. The agents deliberate in multi-turn discussions, drawing on relevant prior experience to aggregate evidence and move toward agreement. The process terminates when agreement is reached or a predefined turn limit is met. A designated coordinator agent then summarizes the discussion, consolidates the accumulated evidence, and outputs the final decision.
To retrieve experience, each agent utterance is first scored using both individual-performance signals and a decayed terminal shared reward. For constructing the experience pool, high-scoring utterances are distilled into textual experiences and added to the pool for subsequent retrieval and integration. 
Experiments show that, on benchmarks spanning medicine, math, and education, MATTRL boosts average performance by 3.67\% over the multi-agent framework and by 8.67\% over comparable single-agent baselines.
Furthermore, we systematically explored multiple credit-assignment schemes for group-credit attribution in experience selection, ranging from naïve shared credit to difference rewards and Shapley-style approximations.
To summarize, our contributions focus on these three perspectives:

\begin{itemize}
    \item We propose the first Multi-agent Test Time Reinforcement Learning framework, MATTRL, leveraging textual experience to enhance the multi-agent system.
    \item We further validate the effect of different credit assignments on experience construction and the final decision.
    \item Experiments conducted on medical, math and education benchmarks achieve a new SOTA performance based on MATTRL.
\end{itemize}

\section{Related Work}

\paragraph{LLM-based multi-agent collaboration.}
Recent advancements in LLM-based multi-agent systems have emphasized scalable collaboration mechanisms for complex task-solving. Surveys \cite{tran2025multi} outline key coordination strategies in LLM-driven multi-agent systems, enabling groups of agents to work collectively at scale. 
MacNet \cite{qian2024scaling} explores the benefits of continuously adding agents to enhance performance in collaborative settings. 
Multi-agent systems utilizing LLMs also emerge as tools for enhancing medical decision-making processes. MDAgents \cite{kim2024mdagents} introduces adaptive collaboration among LLMs to address gaps in clinical reasoning and diagnostics. Multi-agent conversational framework, MAC\cite{chen2025enhancing} boost diagnostic accuracy through interactive agent dialogues. 

\paragraph{Reinforcement learning for LLM reasoning.}
Reinforcement learning techniques have been increasingly applied to refine reasoning capabilities in large language models. Models such as DeepSeek-R1 \cite{guo2025deepseek} demonstrate RL's potential to enhance LLM reasoning without relying on human-annotated data. 
Recent work also systematize RL for reasoning-centric LLMs. SimpleRL-Zoo \cite{zeng2025simplerl} conducts a broad, controlled study of RL on open-base models, showing that careful reward formatting and difficulty curation drive reliable gains across benchmarks. Understanding R1-Zero-Like Training \cite{liu2025understanding} disentangles base-model priors from optimizer effects, identifies length-inducing biases in GRPO, and introduces a debiased variant (Dr.GRPO) that yields strong math results with lightweight recipes. Complementing these, Beyond ``Aha!'' \cite{hu2025beyond} aligns meta-abilities explicitly, spanning deductive, inductive, and abductive skills, via automatically verifiable tasks and targeted RL, achieving consistent improvements over instruction-tuned baselines.

\paragraph{Test-time adaptation and structured experience.}
Test-time adaptation methods allow LLMs to dynamically adjust to new domains during inference without additional training. 
The Test-Time Learning (TTL) paradigm, such as TLM \cite{hu2025test}, adapts models using only unlabeled test data to handle domain shifts effectively. 
Test-Time Reinforcement Learning (TTRL) \cite{zuo2025ttrl} converts test-time scaling signals into pseudo-rewards to train LLMs on unlabeled data, enabling self-evolution and substantial gains.
Study \cite{wang2025far} also  evaluate LLM improvements from structured experience using semantic games as testbeds resistant to saturation.

\paragraph{Credit assignment under collaboration.}
Credit assignment in multi-agent collaborations involving LLMs tackles the challenge of fairly attributing contributions in cooperative settings. LLM-based methods reformulate credit assignment as pattern recognition to achieve efficient and effective distribution in Multi-agent system. Approaches like Shapley-Coop \cite{hua2025shapley} address emergent cooperation in self-interested multi-agent systems through value-based credit allocation. Frameworks such as LLM-MCA \cite{nagpal2025leveraging} utilize large language models for multi-agent credit assignment in reinforcement learning contexts. Systems like CollabUIAgents \cite{headvancing} advance multi-agent learning by incorporating LLM-guided credit re-assignment and synthetic preference data. 

\section{Methodology}

\begin{figure*}
    \centering
    \includegraphics[width=1\linewidth]{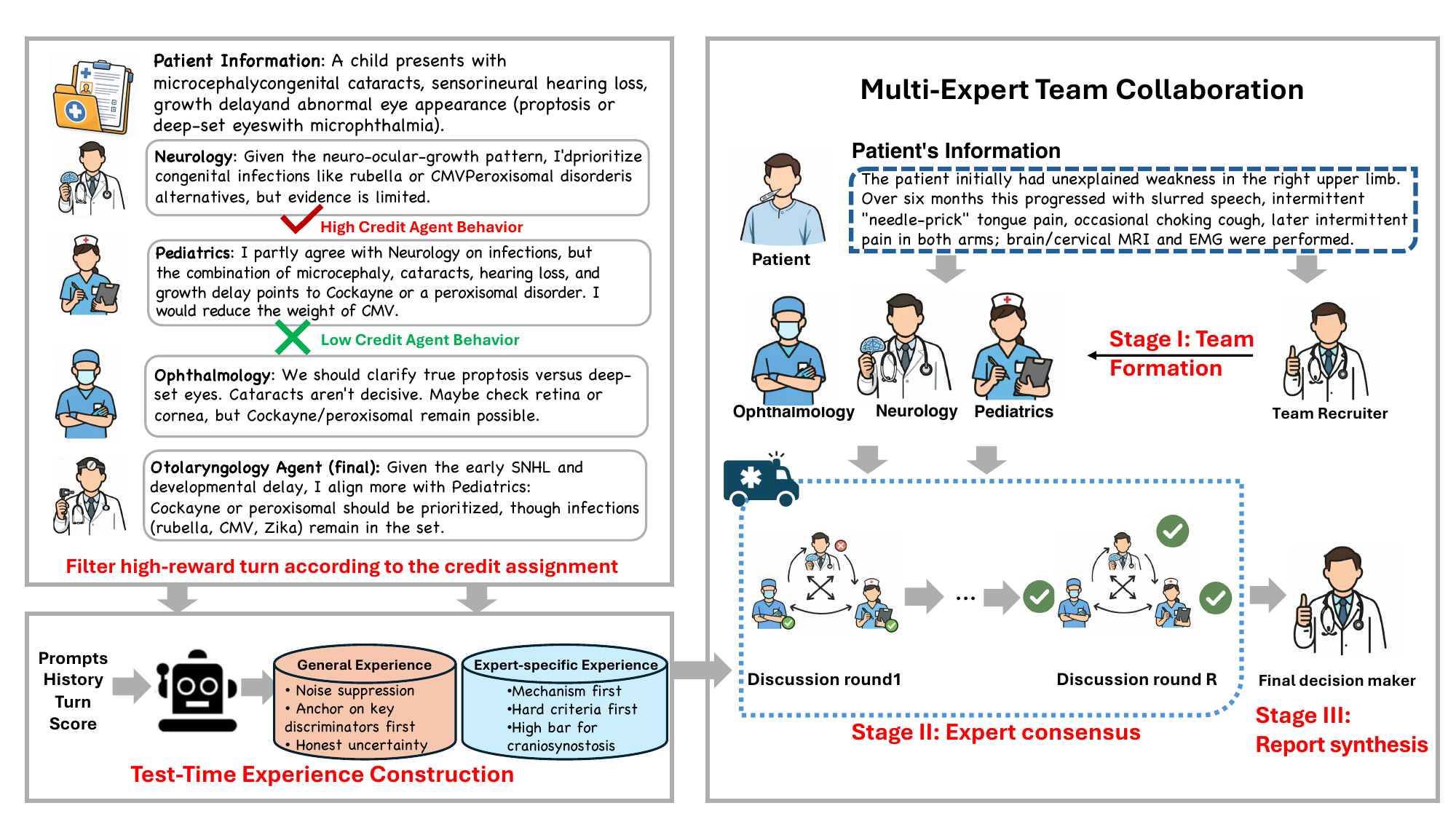}
    \caption{MATTRL overview. The figure uses \textbf{medical diagnosis} as a running example, but the framework is domain-general. \textbf{Math} and \textbf{education} instantiations are in Appendix~\ref{Math-settup} and ~\ref{education-setup}.}
    \label{fig:mattrl_overview}
\end{figure*}

\subsection{Multi-Expert Team Collaboration}
\label{sec:team-collab}

We study a general multi-agent decision-making setting. Each instance provides:
(i) a task record (or user context) $\mathcal{X}$,
(ii) a coordinator agent $\mathrm{LLM}_{\mathrm{Coo}}$,
(iii) an expert catalog $\mathcal{SP}$ (a pool of specialist agents with textual expertise descriptions), and
(iv) a callable test-time experience pool $\mathcal{E}$ (Sec.~\ref{sec:ttec}).
At test time, $\mathrm{LLM}_{\mathrm{Coo}}$ optionally retrieves relevant experiences to strengthen the current decision.
The expert-team consultation follows three stages with a preset maximum of $R_{\max}$ discussion rounds.
Our hospital consultation experiments are a concrete instantiation by interpreting $\mathcal{X}$ as a patient record and $\mathcal{SP}$ as clinical departments.

\paragraph{Stage I: Team formation.}
Rather than letting LLMs freely invent roles, we select an expert team $\mathrm{TEAM}\subseteq \mathcal{SP}$ based on the task record $\mathcal{X}$ using a recruitment prompt (Appendix~\ref{sec:mdt-prompts}) that conditions on $\mathcal{X}$ and each specialist’s expertise description:
\begin{equation}
\mathrm{TEAM} \;\leftarrow\; \mathrm{LLM}_{\mathrm{Coo}}(\mathcal{X}, \mathcal{SP}).
\end{equation}
Each specialist $s\in \mathrm{TEAM}$ maintains a round-indexed opinion set $\mathcal{O}^{(r)}_{s}(\mathcal{X})$ and a convergence flag $f_s^{c}\in\{\mathrm{False},\mathrm{True}\}$ (initialized to $\mathrm{False}$).
We denote the team union at round $r$ as
\begin{equation}
    \mathcal{O}^{(r)}(\mathcal{X}) \;=\; \bigcup_{s\in\mathrm{TEAM}} \mathcal{O}^{(r)}_{s}(\mathcal{X}).
\end{equation}

\paragraph{Stage II: Consensus via experience-augmented dialogue.}
The team proceeds in synchronized rounds $r=0,1,\dots,R_{\max}$.
In each round, each non-converged specialist $s$ retrieves task-relevant experiences and then issues a revised opinion.

We denote the retrieved experience set for $s$ as
\begin{equation}
    \mathrm{ER}_s \;\leftarrow\; \mathrm{Retrieve}\big(\mathcal{E}\,;\, \mathcal{X}, u_s^{(r)}\big),
\end{equation}
where $u_s^{(r)}$ is the current utterance/contextual query formed by specialist $s$ at round $r$.
In our implementation, $\mathrm{Retrieve}(\cdot)$ uses a shared encoder $f(\cdot)$ (\texttt{Qwen3-Embedding-4B}~\cite{zhang2025qwen3}) and a \texttt{FAISS} index~\cite{douze2024faiss} to select top-$K$ entries by cosine similarity. Details are in Appendix~\ref{sec:retrieval_impl}, ~\ref{math-retrieval} and ~\ref{edu-retrieval}.
The retrieved entries are appended to the prompt under a fixed template.

The specialist then updates its opinion conditioned on its previous state and retrieved evidence:
\begin{equation}
    \mathcal{O}^{(r)}_{s}(\mathcal{X})
    \;\leftarrow\;
    \mathrm{LLM}_s\!\big(\mathcal{X},\, \mathcal{O}^{(r-1)}_{s}(\mathcal{X}),\, \mathrm{ER}_s\big).
\end{equation}

We define the incremental update as
\begin{equation}
    \Delta\mathcal{O}^{(r)}_{s}
    \;:=\;
    \mathcal{O}^{(r)}_{s}(\mathcal{X}) \setminus \mathcal{O}^{(r-1)}_{s}(\mathcal{X}).
\end{equation}

Opinions are then synchronized in a meeting step that shares salient updates with all members.
Specifically, $\mathrm{MEETING}(\cdot)$ is a lightweight aggregation operator that takes all specialists' incremental updates $\{\Delta\mathcal{O}^{(r)}_{s}\}_{s\in\mathrm{TEAM}}$ and produces a deduplicated, concise shared bulletin $\Delta\mathcal{O}^{(r)}_{\mathrm{share}}$:
\begin{equation}
    \Delta\mathcal{O}^{(r)}_{\mathrm{share}}
    \;\leftarrow\;
    \mathrm{MEETING}\!\Big(\{\Delta\mathcal{O}^{(r)}_{s}\}_{s\in\mathrm{TEAM}}\Big).
\end{equation}
Each specialist receives $\Delta\mathcal{O}^{(r)}_{\mathrm{share}}$ in the next round's context to align beliefs and avoid redundant discussion.
Each specialist receives $\Delta\mathcal{O}^{(r)}_{\mathrm{share}}$ in the next round’s context.
A specialist is marked converged when no further changes are proposed, i.e., $\Delta\mathcal{O}^{(r)}_{s}=\varnothing$.
The process halts when all specialists converge or when $r=R_{\max}$.

\paragraph{Stage III: Report synthesis and final decision.}
After the bounded discussion, the coordinator agent synthesizes the team’s cumulative evidence into a discussion report $\mathrm{DR}$:
\begin{equation}
    \mathrm{DR}
    \;=\;
    \mathrm{SUMMARY}\!\left[\;\bigcup_{r=0}^{R_{\max}}\;\bigcup_{s\in\mathrm{TEAM}}\; \mathcal{O}^{(r)}_{s}(\mathcal{X})\;\right].
\end{equation}
The coordinator agent may also perform its own retrieval $\mathrm{ER}$ from $\mathcal{E}(\mathcal{X})$, and outputs the final decision $A$ conditioned on the task record and aggregated evidence:
\begin{equation}
    A \;\leftarrow\; \mathrm{LLM}_{\mathrm{Coo}}\!\big(\mathcal{X},\, \mathrm{DR},\, \mathrm{ER}\big).
\end{equation}

\paragraph{Remarks.}
Stage I grounds role selection in a predefined expert catalog $\mathcal{SP}$, Stage II enforces a bounded multi-turn consensus process with explicit convergence checks and retrieval-augmented evidence, and Stage III separates evidence aggregation (report synthesis) from decision making, improving controllability and auditability. 

\subsection{Test-Time Experience Construction}
\label{sec:ttec}

Given a multi-agent transcript with $R$ turns, let $\mathrm{TEAM}$ denote the set of specialist agents.
At turn $t\in\{1,\dots,R\}$, agent $i\in\mathrm{TEAM}$ produces an utterance $u_{i,t}$ under its observable context/history $\mathcal{H}_{i,t}$.
We employ an LLM judge (rubrics in Appendix~\ref{rubrics}, ~\ref{app:math:utterance_scoring}, and ~\ref{utterance_scoring}) to evaluate each utterance along domain-relevant axes (e.g., correctness, information gain, relevance to the task, clarity, \emph{etc.}), yielding an \emph{individual score}:
\begin{equation}
    s_{i,t}
\;=\;
\phi_{\text{LLM}}\!\big(u_{i,t},\, \mathcal{H}_{i,t};\, \text{Rubric}\big)
\in[0,1].
\end{equation}

\paragraph{Contribution ratio and terminal shared reward.}

Assume we obtain a single \emph{terminal} team-level outcome score $G$ at the end of the consultation
(e.g., task success), where $G\in[0,1]$.
Let $R$ be the actual number of turns (with $R\le R_{\max}$).
We allocate $G$ back to each turn via a decay kernel and split each turn’s share across agents by contribution ratios. Define per-turn decay weights
\begin{equation}
    w_t \;=\; \gamma^{\,R-t}
\end{equation}
The later turns receive higher weight when $\gamma<1$.
Each agent’s contribution ratio $c_{i,t}$ is estimated by proportional normalization of per-agent scores within each turn:
\begin{equation}
    c_{i,t}=\frac{s_{i,t}}{\sum_{j\in\mathrm{TEAM}} s_{j,t} + \epsilon},\qquad s_{i,t}\ge 0,
\end{equation}
where $\epsilon$ avoids division by zero.

\paragraph{Turn-level reward for each agent.}
We fuse individual and terminal team signals:
\begin{equation}
\label{eq:step-reward-terminal}
r_{i,t}
\;=\;
\lambda\, s_{i,t}
\;+\;
(1-\lambda)\, G \cdot w_t \cdot c_{i,t},
\qquad
\lambda\in[0,1].
\end{equation}

\paragraph{Selection of high-value utterances.}
To construct reusable test-time experiences, we select high-value snippets using a threshold:
\begin{equation}
\mathcal{I}_i^{\text{keep}}
=
\big\{\, t \;\big|\; r_{i,t}\ge \tau \,\big\}.
\end{equation}

\paragraph{From high-scoring utterances to textual experience.}
For each $(i,t)\in\mathcal{I}_i^{\text{keep}}$, we map the context $\mathcal{H}_{i,t}$, utterance $u_{i,t}$, and quantitative signals $r_{i,t}$ into a structured, retrievable \emph{textual experience entry} using an LLM summarizer (prompt templates in Appendix~\ref{summarizer}):
\begin{equation}
\label{eq:exp-map}
e_{i,t}
\;=\;
\Psi_{\text{LLM}}\!\Big(\mathcal{H}_{i,t},\, u_{i,t},\, r_{i,t};\, \text{Template}_{\text{exp}}\Big).
\end{equation}

This yields a test-time experience pool
\begin{equation}
    \mathcal{E}
\;=\;
\big\{\, e_{i,t} \;\big|\; i\in\mathrm{TEAM},\; t\in\mathcal{I}_i^{\text{keep}} \,\big\},
\end{equation}
We define a \emph{textual experience entry} as a compact, structured text record that is easy to retrieve and reuse. Each entry stores (i) minimal task context for retrieval, (ii) the actionable step taken, and (iii) a short rationale for the assigned credit.

\section{Experiments}

\begin{table*}[ht]
    \centering
    \vspace{-2mm}
    \begin{tabular}{c|ccccc}
    \hline
        Method & Hit@1 & Hit@3 &Hit@5 & Hit@10 & MRR \\
        \hline
        MDAgent                 & 0.32 & 0.49 & 0.57 & 0.68 & 0.46 \\
        RareAgents      & 0.29 & 0.38 & 0.47 & 0.68 & 0.42 \\
        RareAgent-Refined      & 0.35 & 0.49 & 0.57 & 0.70 & 0.47 \\
        MATTRL  & \textbf{0.39} & \textbf{0.51} & \textbf{0.61} & \textbf{0.75} & \textbf{0.51}\\ \hline
    \end{tabular}
    \vspace{-2mm}
    \caption{Experimental Results on Baselines and MATTRL for medicine benchmark}
    \vspace{-2mm}
    \label{overall performance}
\end{table*}

\subsection{Setup}

\paragraph{Datasets and Domain Settings}
In \textbf{Medicine} setting, RareBench~\cite{chen2024rarebench} evaluates LLMs as rare-disease specialists across four tasks. We focus on Task 4 (differential diagnosis among universal rare diseases) with 2,185 cases covering 421 diseases, and cast the task as a multi-agent consultation: an attending agent orchestrates domain specialists to independently propose and justify differential diagnoses from the patient record, critique peers’ evidence, and iteratively refine toward a consensus shortlist.
\textbf{Math}: We utilize HLE (Humanity’s Last Exam)~\cite{phan2025humanity} with text-only math problems (856 samples), a challenging benchmark of expert-level questions, to assess collaborative problem solving. We report exact-match solve rate via LLM judgement and quantify the improvement brought by multi-agent deliberation with test time experience. 
\textbf{Education}:We study teaching-oriented interaction with a three-stage designs: pre-test, instruction, and post-test. The student first answers with reasoning. Then a teacher, given the question, gold answer, and the student’s response, conducts a two-round teaching dialogue. Finally, the student re-answers. We sample 300 questions from SuperGPQA~\cite{du2025supergpqa} with GPT-4o as the student and GPT-5 as the teacher, and measure learning gains by post-test accuracy improvement. We also demonstrate the detailed examples, settings and prompts for these three domains in Appendix~\ref{APP-Medicine}, ~\ref{APP-Mathematics} and ~\ref{APP-Education}.

\paragraph{Baselines.}
In \textbf{medicine} settings, We compare against two agentic baselines. \textit{MDAgents}~\cite{kim2024mdagents} is an adaptive collaboration framework that estimates case complexity, recruits an appropriate team, performs multi-turn analysis–synthesis, and ends with moderator review. Its dynamic structure and moderation/knowledge components improve medical QA and diagnosis. RareAgents~\cite{chen2024rareagents} targets rare-disease diagnosis via a patient-centered Multi-disciplinary Team (MDT) with specialist orchetration, case-memory retrieval, and tool use. Since its memory corpus and tool library are not released, we reimplement the MDT-only version. We also introduce \textit{RareAgents-Refined}, a prompt-engineered variant that enforces role-focused, critical peer review and discourages fabricated tests/results, reducing confirmation bias and hallucinations and yielding consistent gains. 
For \textbf{math} and \textbf{education} domains, we use a \textbf{single-agent} solver/teacher that directly performs the task as one baseline. We then compare it against our \textbf{multi-agent} instantiation described in Section~\ref{sec:team-collab}, where multiple experts independently propose, critique, and iteratively refine solutions (or teaching moves) with periodic synchronization/aggregation. This isolates the effect of test-time experience.


\paragraph{Metrics}
\textbf{Medicine.} We report \textit{Hit@k} and \textit{MRR} on the attending agent’s \emph{final ranked differential list/shortlist}, where Hit@k is the fraction of cases whose ground-truth disease appears within the top-$k$ predictions, and MRR averages $1/\text{rank}$ of the correct disease. Higher is better.
\textbf{Math.} We report exact-match solve rate (\textit{Acc}), where a problem is counted as solved if the final answer matches the reference under an LLM judge. 
\textbf{Education (SuperGPQA).} We measure learning by pre-test and post-test accuracy and report learning gains as $\Delta \textit{Acc}=\textit{Acc}_{\text{post}}-\textit{Acc}_{\text{pre}}$ (higher indicates stronger instructional improvement).

\paragraph{Paremeters Settings}

We use GPT-5 \cite{openai2025gpt5} as the backbone model is our MATTRL framework and other aforementioned LLMs are also GPT-5. The number of experts is 3, and the maximum conversation turns are limited to 3.
For experience text construction, we select 30 cases. For all utterance from agents, we extract the Top 25\% scored records for further construction.

\subsection{Results}

As demonstrated in Table~\ref{overall performance}, in medicine task, MATTRL achieves the strongest overall retrieval quality. Averaged over k = 1, 3, 5, and 10, its Hit@k is 0.565, higher than MDAgent at 0.515 and RareAgents-Refined at 0.528, and it also attains the highest MRR of 0.51. The most pronounced advantages appear at Hit@1, indicating better top-rank precision, and at Hit@10, indicating more reliable shortlist coverage. Overall, the results suggest that test-time collaborative adaptation yields benefits beyond those achievable through prompt optimization alone.

As shown in Table~\ref{tab:math}, the single-agent baseline achieves an exact-match accuracy of 0.27 on HLE. Introducing multi-agent deliberation improves performance to 0.33, indicating a modest benefit from parallel proposal and critique. MATTRL yields a larger gain, reaching 0.36, suggesting that test-time experience further strengthens collaborative problem solving beyond deliberation alone.

For Education, as shown in Table~\ref{education}, all methods start from the same pre-test accuracy (\(\textit{Acc}_{\text{pre}}=0.44\)), ensuring a controlled comparison where improvements reflect instructional effectiveness rather than initial student performance. The single-agent teacher increases accuracy to \(\textit{Acc}_{\text{post}}=0.60\)
(\(\Delta\textit{Acc}=0.16\)). Replacing it with a multi-agent teacher that proposes and critiques teaching moves yields a much larger gain, suggesting that deliberation helps identify misconceptions and select more effective explanations. MATTRL further achieves the best post-test performance at \(\textit{Acc}_{\text{post}}=0.77\) with the highest learning gain (\(\Delta\textit{Acc}=0.33\)), nearly doubling the improvement of the single-agent baseline. Overall, the results indicate that collaboration substantially enhances teaching outcomes, and test-time experience provides additional benefits beyond collaboration alone.


\begin{table}[t]
    \centering
    \vspace{-1mm}
    \setlength{\tabcolsep}{4pt}
    \renewcommand{\arraystretch}{1.15}
    \newcommand{\scoredelta}[2]{#1\raisebox{-0.6ex}{\scriptsize\,\textcolor{black}{#2}}}
    \newcommand{\scoredeltaNeg}[2]{#1\raisebox{-0.6ex}{\scriptsize\,\textcolor{black}{#2}}}

    \begin{tabular}{l|ccc}
        \hline
        Method 
        & Single Agent & Multi-Agent & MATTRL \\
        \hline
        \textit{Acc} &
        0.27 &
        \scoredelta{0.33}{(+0.06)} &
        \scoredeltaNeg{0.36}{(+0.09)} \\
        \hline
    \end{tabular}
    \vspace{-2mm}
    \caption{\textbf{Math (Accuracy Comparison with Per-Method Improvement).} We report exact-match accuracy on HLE math problems. Numbers in the bottom-right indicate the absolute change in accuracy relative to the single agent baseline}
    \vspace{-4mm}
    \label{tab:math}
\end{table}
\vspace{-3mm}



\begin{table}[ht]
    \centering
    \vspace{-1mm}
    \begin{tabular}{c|ccc}
    \hline
        Method & $\mathit{Acc}_{\text{pre}}$ & $\mathit{Acc}_{\text{post}}$ & $\Delta \mathit{Acc}$ \\
    \hline
        Single Agent & 0.44 & 0.60 & 0.16 \\
        Multi-Agent  & 0.44 & 0.73 & 0.29 \\
        MATTRL       & 0.44 & 0.77 & 0.33 \\
    \hline
    \end{tabular}

    \vspace{-2mm}
    \caption{\textbf{Education (Learning Gains in a Pre-test $\rightarrow$ Tutoring $\rightarrow$ Post-test Setup).}
    We report pre-test accuracy ($\mathit{Acc}_{\text{pre}}$), post-test accuracy ($\mathit{Acc}_{\text{post}}$),
    and learning gain ($\Delta\mathit{Acc}=\mathit{Acc}_{\text{post}}-\mathit{Acc}_{\text{pre}}$).}
    \label{education}
\end{table}
\vspace{-3mm}

\section{Analysis}

\textbf{All ablations and analysis conducted below are based on medicien dataset (RareBench)}.

\subsection{Group-to-Agent Credit Assignment}
We compare naive averaging, Difference Rewards, and Shapley-style approximations for attributing team returns at each turn to individual agents. 

As we mentioned in section~\ref{sec:ttec}, We compute agent-specific \emph{credit scores} $q_{i,t}$ for agent $i$ at turn $t$ and map them to contribution ratios via a shared normalization to ensure comparability:
\begin{equation}
c_{i,t} \;=\; \frac{\exp(\beta\, q_{i,t})}{\sum_{j\in\mathrm{TEAM}}\exp(\beta\, q_{j,t})} \qquad \beta>0.
\end{equation}

\textbf{Difference Rewards.}
For agent $i$ at turn $t$, define the counterfactual where $i$ is neutralized while others remain:
\begin{equation}
q^{\mathrm{Diff}}_{i,t} \;=\; F_t(\mathrm{TEAM}) - F_t(\mathrm{TEAM}\setminus\{i\})
\end{equation}
where $F_t(\cdot)$ is the turn-$t$ team objective (e.g., consensus gain or hypothesis-space reduction). In practice, $F_t(\mathrm{MDT}\setminus\{i\})$ is approximated by rerunning the turn with $i$’s utterance replaced by a no-op, or via a learned proxy (Appendix).

\textbf{Shapley-style approximations.}
The Shapley value averages $i$’s marginal effect across orders:
\begin{equation}
q^{\mathrm{Shap}}_{i,t} \;=\;
\mathbb{E}_{\pi}\!\left[
F_t\!\big(S^{<i}_\pi \cup \{i\}\big) - F_t\!\big(S^{<i}_\pi\big)
\right]
\end{equation}

with $S^{<i}_\pi$ the set of agents preceding $i$ in permutation $\pi$. We estimate $q^{\mathrm{Shap}}_{i,t}$ via $K$ Monte Carlo permutations (or small-coalition sampling) with cached $F_t(\cdot)$ to control cost. Unless stated otherwise, all schemes use the same $F_t(\cdot)$ and the same normalization (identical $\beta$) before feeding $c_{i,t}$ into the decay-weighted terminal allocation in \textit{Contribution ratio and terminal shared reward.}

\begin{table}[ht]
    \centering

    \vspace{-1mm}
    \begin{tabular}{c|ccccc}
    \hline
        Method & Hit@1 & Hit@3 &Hit@5 & Hit@10 \\
        \hline
        Naive &0.39   &0.51	&0.61	&0.75	 \\
        Difference  &0.40  &0.53    &0.61   &0.74   \\
        Shapley  &0.35	&0.49	&0.59	&0.75	 \\ \hline
    \end{tabular}
    \vspace{-2mm}
    \caption{Performance comparison among different credit assignments for experience construction. Naive represents the Naive method we mentioned in section~\ref{sec:ttec}, Difference denotes the Difference Rewards and Shapley is Shapley-style approximations.}
    \label{credit assignment}
\end{table}
\vspace{-3mm}




As shown in Table~\ref{credit assignment}, \textsc{Difference} yields the best strict-precision
performance (Hit@1/3 = 0.40/0.53), outperforming \textsc{Naive} (0.39/0.51) and \textsc{Shapley} (0.35/0.49). At broader cutoffs the methods are similar: Hit@5 is tied for \textsc{Difference}/\textsc{Naive} (0.61) and Hit@10 is nearly identical (0.74--0.75). We attribute \textsc{Difference}'s gains on tight metrics to reduced free-riding noise: contrasting the full team with a counterfactual where agent \(i\) is neutralized better isolates decisive turns and produces sharper credit peaks after normalization. By contrast, \textsc{Shapley} tends to spread credit across coalitions (and is variance-prone under limited permutations), which dilutes peaks and hurts Hit@1/3 despite comparable Hit@10.

\paragraph{Why Shapley underperforms.} We observe that Shapley-style selection tends to reward \textit{peer-review/alignment behaviors} that improve coherence and consensus but have limited influence on the decisive inference steps. Since Shapley averages marginal effects across many coalitions, sharp decision moves are diluted while low-variance meta-behaviors accumulate steady credit (e.g., “integrate peer comments coherently,” “maintain cross-specialty consensus”). By contrast, Naive more often elevates \textit{decision-centric hints} with short feedback loops because it ties credit to single-run outcome deltas (e.g., “prioritize MMA over PA when biomarkers dominate,” “merge weakly anchored subtypes into a low-priority bucket”), yielding sharper hypothesis ranking and stronger top-rank accuracy. Beyond hit rates, compute and stability also favor Difference. Shapley needs many marginal evaluations and has higher estimator variance unless heavily sampled; Naive is cheapest but sensitive to correlated noise. Difference offers a practical middle ground with one counterfactual per agent, providing a low-variance, high-leverage signal at modest cost. Overall, we recommend Shapley when fairness is paramount and budget allows, Naive as a low-cost baseline, and Difference as the default when precision and efficiency matter.







\subsection{Adaptive collaboration between single agent and multi-agent framework}

To further improve the practicality of MATTRL, we additionally compare against a single-agent baseline using chain-of-thought (CoT) reasoning and develop an Adaptive method that learns to route each case to either the single agent or MATTRL. The classifier makes the routing decision based on features capturing symptom complexity, need for multidisciplinary consultation, number of specialties involved, cross-specialty divergence, and risk of single-expert misguidance. As shown in Table~\ref{adaptive}, the single-agent CoT baseline is already strong, and the Adaptive router further improves performance, achieving average gains of 10\% over the single agent and 5.5\% over MATTRL.

\begingroup
\setlength{\tabcolsep}{4pt}   
\begin{table}[ht]

    \centering
    \vspace{-2mm}
    \begin{tabular}{c|ccccc}
    \hline
        Method & Hit@1 & Hit@3 &Hit@5 & Hit@10 \\
        \hline
        Single-Agent &0.39   &0.49	&0.56	&0.64	 \\
        MATTRL  &0.39  &0.51    &0.61   &0.75   \\
        Adaptive  &\textbf{0.45}	&\textbf{0.58}	&\textbf{0.66}	&\textbf{0.79}	 \\ \hline
    \end{tabular}
    \caption{Results of Single-Agent, MATTRL, and Adaptive Router (Adaptive in below table).}
    
    \label{adaptive}
\end{table}
\endgroup
\vspace{-3mm}

Single-agent excels when cases show standardized diagnostic “fingerprints” that a one-shot integration can resolve, evidence is concentrated in one specialty, and the task prioritizes internal consistency with a concise explanation. Multi-agent is stronger when evidence spans multiple specialties or modalities and needs cross-validation, the goal extends to risk assessment/care planning/test prioritization, and the task benefits from systematic counterfactuals and competing hypotheses for robust differentials. This aligns with our analysis for the classifier in adaptive method and the error analysis for both single agent and MATTRL.

Our classifier routed 282 cases to the single-agent solver and 840 to MATTRL. Empirically, many instances that are internally consistent are solvable by the single agent, yet the multi-agent discussion can introduce noise that harms accuracy on those same cases. A Venn-style breakdown of correctness shows: Only the single agent solves around 300 cases, only MATTRL solves 400+ cases, and both solve 357 cases.

\subsection{Scaling with Team Size}

\begin{figure}[t]
\centering
\begin{tikzpicture}
\begin{axis}[
    width=\linewidth, height=6.5cm,      
    xlabel={Team size (number of experts)},
    ylabel={Accuracy (\%)},
    label style={font=\small},            
    tick label style={font=\footnotesize,/pgf/number format/fixed},  
    xmin=1, xmax=9,
    xtick={1,3,5,7,9},
    ymin=30, ymax=90,
    ymajorgrids, xmajorgrids,
    line width=1pt,
    mark size=2.5pt,
    legend style={
        font=\footnotesize,               
        draw=none, fill=none,
        at={(0.5,-0.18)},                 
        anchor=north,
        legend columns=4,                 
        /tikz/every even column/.append style={column sep=8pt}
    },
]

\addplot+[mark=*] coordinates {(1,37) (3,39) (5,37) (7,35) (9,38)};
\addlegendentry{Hit@1}

\addplot+[mark=square*] coordinates {(1,50) (3,50) (5,49) (7,50) (9,52)};
\addlegendentry{Hit@3}

\addplot+[mark=triangle*] coordinates {(1,59) (3,61) (5,61) (7,61) (9,63)};
\addlegendentry{Hit@5}

\addplot+[mark=diamond*] coordinates {(1,72) (3,75) (5,86) (7,84) (9,82)};
\addlegendentry{Hit@10}

\end{axis}
\end{tikzpicture}
\vspace{-6mm}
\caption{GPT-5 Multi-Agent: Acc. by Team Size.}
\label{team size}
\vspace{-2mm}
\end{figure}
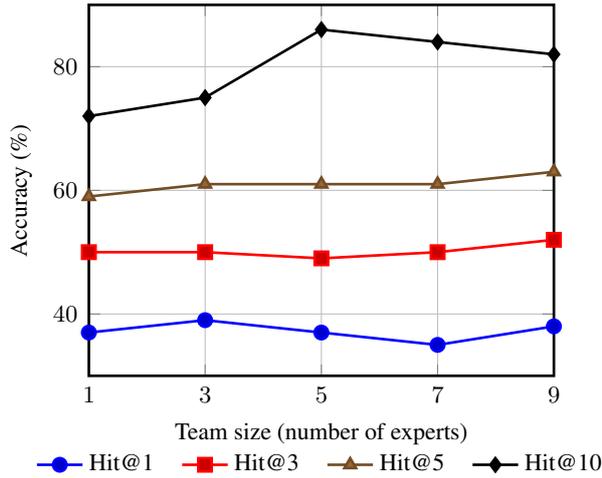

We study how performance scales as the number of collaborating agents increases (e.g., 1, 3, 7, 9). As shown in Figure~\ref{team size}, increasing the number of agents does not uniformly improve performance. For Hit@1, accuracy peaks at three agents and then declines as the team grows. Because Hit@1 requires strict precision, larger teams introduce more divergent opinions and make consensus harder to reach. In contrast, Hit@3 and Hit@5 exhibit modest, steady gains with scale. Hit@10 benefits the most from scaling, as broader discussions surface more plausible candidates and are more tolerant to noise. Notably, a three-agent team outperforms a single agent by about 14\% on Hit@10. Practically, smaller teams (e.g., three agents) are preferable for high-precision decisions, whereas larger teams help when broader recall is desired.

\begin{figure}[ht]
\centering
\begin{tikzpicture}
    \node[mybox, inner sep=3pt] (box){%
        \begin{minipage}{0.98\columnwidth} 
        \scriptsize 
        \setstretch{1.05} 
        \setlength{\parskip}{0pt}
        \setlength{\parindent}{0pt}

        \begin{flushleft}
        \vspace{2mm}
        \textbf{General Experience: }
        
        \includegraphics[width=0.28cm]{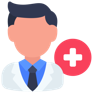}\hspace{4pt}\textbf{Noise suppression}: Oppose the vague “as long as it matches” alignment. Broad statements like “can fit multiple diseases / seems consistent” dilute discriminative power and prolong an unhelpful tail. The requirement is to clearly explain the mechanistic consistency from feature → diagnosis; otherwise, down-weight or exclude it.\\
        \includegraphics[width=0.28cm]{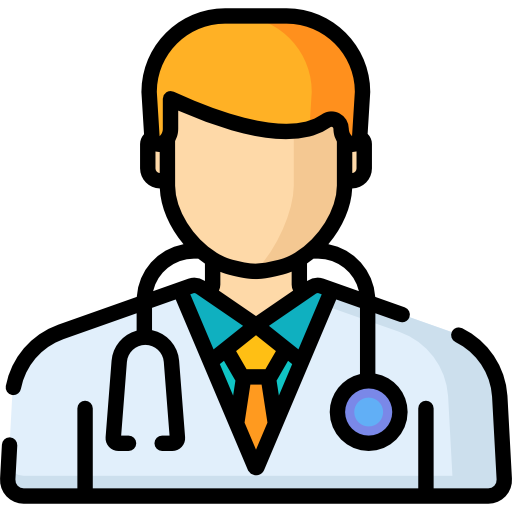}\hspace{4pt}\textbf{Anchor on key discriminators first}: Use the few features that “most sharply split candidate diseases in one cut” and are observed across multiple specialties as the main backbone (main sequence) for ranking. Other evidence should only serve as fine-tuning, not overturn the backbone.\\
        \includegraphics[width=0.28cm]{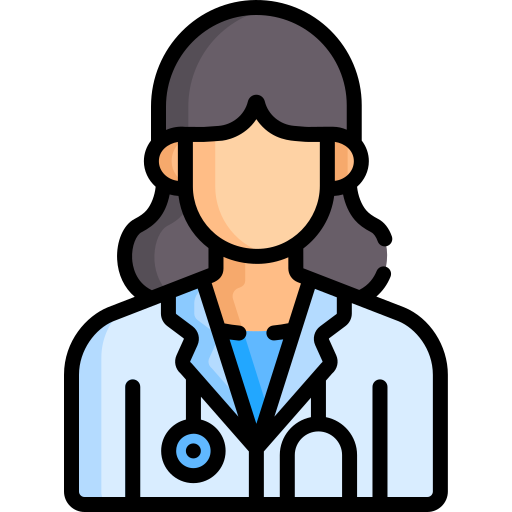}\hspace{4pt}\textbf{Honest uncertainty}: When the evidence is insufficient to make a unique choice, explicitly state “insufficient to select a single diagnosis” to avoid over-commitment; ranking must stay constrained by evidence.\\

        \vspace{2mm}
        \textbf{Disease-specific Experience: }

        \includegraphics[width=0.28cm]{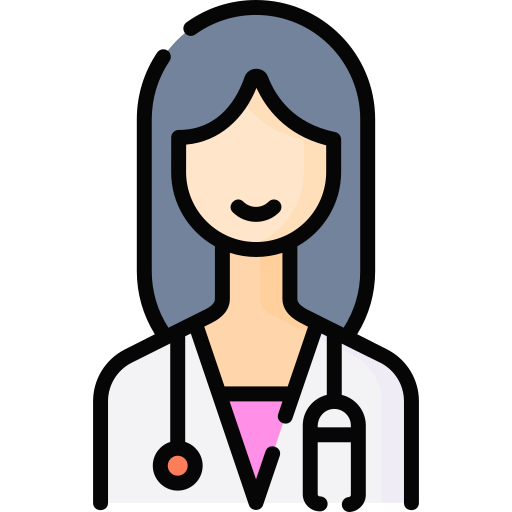}\hspace{4pt}\textbf{Mechanistic clarification before assumptions}: For the cause of leukocoria (“white pupil”), attribution (lens vs retina) must be clarified first; if evidence is lacking, state “insufficient for attribution.” This prevents prematurely over-weighting a certain subtype/spectrum.\\
        \includegraphics[width=0.28cm]{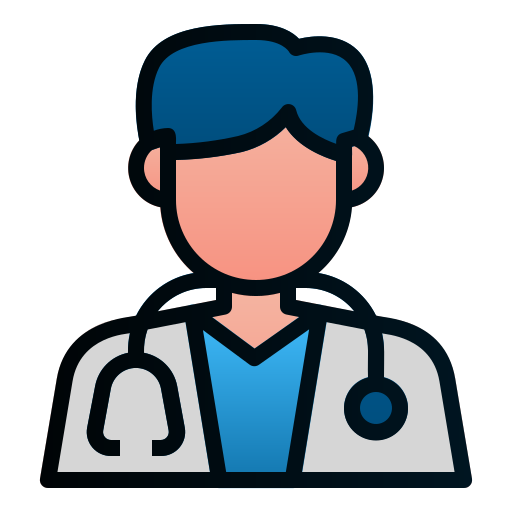}\hspace{4pt}\textbf{Let hard criteria dominate near-neighbor ordering}: For axial/epiphyseal hard features like odontoid hypoplasia or delayed ossification, require an explicit explanation of how they shift the relative positions of close candidates. Within related spectra, use these high-weight hard indicators as tie-breakers, rather than vague impressions.\\
        \includegraphics[width=0.28cm]{assets/doctor1.png}\hspace{4pt}\textbf{Set a high evidence threshold for the craniosynostosis spectrum}: Non-specific morphologies such as shallow orbits/proptosis, thick skull, etc., are insufficient to raise its weight. Without hallmark evidence of suture involvement, keep it low-ranked and state this explicitly. Also, remind to explicitly address cranial signs that contradict the main hypothesis, to avoid anchoring bias.\\
        \end{flushleft}
        \end{minipage}
    };

    \node[fancytitle, rounded corners, right=8pt] at (box.north west) {Experiences};
\end{tikzpicture}
\vspace{-6mm}
\caption{General \& disease-specific experience}
\label{experience}
\end{figure}
\vspace{-3mm}
\subsection{Experience Examples}
Figure~\ref{experience} shows two kinds of reusable test-time experiences that MATTRL extracts from consultation transcripts. \emph{General experiences} are cross-disease rules that improve discriminability and keep discussion disciplined. For instance, they require mechanism-grounded justifications instead of vague ``seems consistent'', prioritize a small set of high-yield discriminators as the ranking backbone, and state uncertainty explicitly when evidence is weak. \emph{Disease-specific experiences} are concise, concrete checks that guide fine-grained ordering among close candidates (e.g., first clarify the locus of leukocoria before assuming a subtype; let high-weight skeletal markers adjust relative ranks; keep craniosynostosis low without direct evidence of suture involvement). Practically, we select utterances with higher reward via credit assignment, distill their underlying rationale into brief, textual experience snippets, and retrieve them at inference to stabilize multi-agent deliberation and improve accuracy without updating model weights.

\subsection{Few-shot vs. Test-time Experience}

To test whether MATTRL’s gains stem merely from supplying extra context, we compare MATTRL with RareAgents augmented by few-shot exemplars (containing patient information and the final diagnosis). For each test case, 3 random exemplars are prepended to the conversation. As shown in Table~\ref{fewshot}, few-shot prompting yields only a minor improvement in Hit@1 while reducing Hit@3/5/10. This indicates that MATTRL’s advantage arises from its structured experience integration rather than from simply adding more information.

\begingroup
\setlength{\tabcolsep}{5.5pt}    

\begin{table}[ht]
    \centering
    \vspace{-2mm}
    \begin{tabular}{@{}c@{}|@{}cccc@{}}
    \hline
        Method & Hit@1 & Hit@3 & Hit@5 & Hit@10 \\
        \hline
        RareAgents   & 0.35 & 0.49 & 0.57 & 0.70 \\
        + Fewshot  & 0.37 & 0.48 & 0.55 & 0.68 \\
        \hline
        MATTRL       & 0.39 & 0.51 & 0.61 & 0.75 \\
        \hline
    \end{tabular}
    
    \caption{Comparison with fewshot learning, where we add 3 examples at the beginning of each conversation.}
    \label{fewshot}
    \vspace{-4mm}
\end{table}
\endgroup

\vspace{-4mm}

\section{Conclusion}

We introduced \textbf{MATTRL}, a test-time adaptation framework that strengthens multi-agent reasoning by injecting \emph{structured textual experience} into deliberation. 
MATTRL builds a small expert team, curates an experience pool from high-value dialogue turns via group-to-agent credit assignment, and retrieves these experiences to guide subsequent collaboration. 
Across \textbf{medicine}, \textbf{math}, and \textbf{education}, it consistently outperforms single- and multi-agent baselines, showing that experience-conditioned collaboration improves robustness under distribution shift.
We further analyzed credit-assignment strategies and find that \textsc{Difference} rewards provide a strong accuracy and efficiency trade-off for experience construction. Finally, an adaptive router that selects between single-agent inference and MATTRL yields additional gains by matching collaboration style to case complexity.

\section*{Limitations}

We recognize two practical limitations remain. First, the method’s inference-time compute and latency grow with multi-agent rollouts and exploration budget. 
Second, a continually growing test-time experience pool is vulnerable to drift: stale, duplicated, or spurious heuristics may accumulate. 
Looking ahead, we will (i) introduce dynamic budget controllers and confidence-based early stopping to cap cost without hurting accuracy, and (ii) add lifecycle management for experiences (recency weighting, de-duplication, anomaly screening) to preserve precision over time.

\bibliography{custom}

\appendix
\clearpage

\section{Medicine}
\label{APP-Medicine}

\subsection{Detailed Setup}
\label{app:med:detailed_setup}

\paragraph{Task and data (RareBench Task 4).}
We instantiate MATTRL as an MDT-style workflow for rare-disease differential diagnosis on RareBench Task 4~\cite{chen2024rarebench}.
Each instance provides a patient record $\mathcal{X}$ and the system outputs a ranked top-10 differential list.
We evaluate with Hit@k and MRR as defined in the main text.

\paragraph{Agents, specialist pool, and recruitment.}
The system consists of a coordinator/chair agent $\mathrm{LLM}_{\mathrm{Coo}}$ and a predefined specialist catalog $\mathcal{SP}$ (Appendix~\ref{app:med:specialist_pool}).
$\mathrm{LLM}_{\mathrm{Coo}}$ recruits a small MDT $\mathrm{TEAM}\subseteq\mathcal{SP}$ using the recruitment prompt (Appendix~\ref{sec:mdt-prompts}),
grounding role selection in real clinical departments rather than free-form role invention.

\paragraph{MDT interaction protocol and prompts.}
Given $\mathrm{TEAM}$, specialists follow role-specific opinion prompts and produce a strict top-10 list each round (Appendix~\ref{sec:mdt-prompts}).
We run synchronized multi-round discussion with a maximum of $R_{\max}$ rounds as described in Sec.~\ref{sec:team-collab}.
The chair then synthesizes a discussion report and outputs the final ranked list using the final-decision prompt (Appendix~\ref{sec:mdt-prompts}).
Experience-augmented prompting uses the standardized injection template in Appendix~\ref{sec:exp-aug-prompt}.

\paragraph{Utterance scoring and judge rubric.}
To score specialist utterances for experience construction, we use an LLM judge with the rubric defined in Appendix~\ref{rubrics},
producing per-utterance scores $s_{i,t}\in[0,1]$ (Eq.~(9) in the main text).
These individual scores are combined with a terminal case-level outcome signal via the decay-weighted allocation (Eq.~(10)--(12) in the main text).

\paragraph{Experience extraction and summarization.}
High-scoring utterances are distilled into structured textual experiences using an LLM summarizer with the template in Appendix~\ref{summarizer}.
Each entry follows the \texttt{ACTION}/\texttt{EXPERIENCE} schema used in our experience-augmented prompt template (Appendix~\ref{sec:exp-aug-prompt}).

\paragraph{Indexing and retrieval.}
At test time, each specialist retrieves relevant experience entries conditioned on the case and round context.
We detail the embedding model, similarity metric, and top-$K$ retrieval procedure in Appendix~\ref{app:med:retrieval}.
Retrieved experiences are appended to prompts via the injection block in Appendix~\ref{sec:exp-aug-prompt}, keeping model weights fixed while providing dense guidance.

\subsection{Description of Specialist Pool}
\label{app:med:specialist_pool}

This pool covers core inpatient and outpatient specialties frequently involved in complex differential diagnosis. It is designed to balance breadth with depth, enabling targeted and efficient MDT assembly.

\begin{table}[h!]
\centering
\resizebox{0.45\textwidth}{!}{%
\begin{tabular}{ll}
\toprule
Pediatrics & Urology \\
Hematology & Rheumatology \\
Psychiatry & Pulmonology \\
Dentistry & Endocrinology \\
Allergy and Immunology & Cardiology \\
Pathology & Neurology \\
Obstetrics and Gynecology & Ophthalmology \\
Dermatology & Geriatrics \\
Traditional Chinese Medicine & Nephrology \\
Oncology & General Practice \\
Gastroenterology & Infectious Diseases \\
Rehabilitation Medicine & Otorhinolaryngology \\
\bottomrule
\end{tabular}%
}
\caption{List of 24 Departments from Specialist Pool.}
\label{tab:specialist_departments_2col}
\end{table}



\subsection{Experience-Augmented Prompt Template}
\label{sec:exp-aug-prompt}
This template integrates retrieved experience into the base diagnostic instruction. The \emph{Experience Context} block is formatted to remain model-friendly while improving calibration and coverage of edge patterns.

\begin{promptbox}{Experience-Augmented Diagnostic Prompt}
\small\setstretch{1.2}\RaggedRight
You are an expert clinician AI agent participating in a diagnostic reasoning task.\\[2pt]
Your goal is to reason step by step and propose the top 10 possible diagnoses for the given case.\\[6pt]

\textbf{Patient Case}\\
\code{\{question\}}\\[6pt]

\textbf{Experience Context}\\
Insert domain experience as structured hints; each item pairs an \code{ACTION} with an \code{EXPERIENCE}. The block is optional; omit if empty.
\begin{lstlisting}[style=wraptt]
===== EXPERIENCE HINTS =====
- ACTION: <action_key_1>
  EXPERIENCE: <concise, actionable rule or prior finding>
- ACTION: <action_key_2>
  EXPERIENCE: <concise, actionable rule or prior finding>
...
===== END OF EXPERIENCE HINTS =====
\end{lstlisting}

\textbf{Output Requirements}\\
Return exactly the following structure; no extra prose outside it.
\begin{lstlisting}[style=wraptt]
1) Reflection (2-3 sentences)
2) <diagnosis> block with exactly 10 numbered items:
   1. [Diagnosis 1]: [1-2 sentence rationale]
   ...
   10. [Diagnosis 10]: [1-2 sentence rationale]
</diagnosis>
\end{lstlisting}

\textbf{Hard Constraints}
\begin{itemize}
  \item Base reasoning strictly on the Patient Case and the Experience Context (if provided).
  \item Do not invent external facts, tests, or treatments.
  \item Do not copy reasoning from other roles (if used in a multi-agent setting).
  \item If key information is missing, write ``insufficient evidence'' instead of guessing.
\end{itemize}
\end{promptbox}

\subsection{Prompts for Multi-disciplinary Team Collaboration}
\label{sec:mdt-prompts}
These prompts orchestrate role selection, role-specific reasoning, and peer oversight. The design favors minimal, structured outputs to simplify downstream aggregation and evaluation.

\begin{promptbox}{MDT Recruitment Prompt}
\small\setstretch{1.2}\RaggedRight
You are the Chief Medical Officer assembling a single MDT.\\[2pt]
\textbf{Case:} \code{\{question\}}\\[2pt]
From this specialist pool: \code{\{POOL\}}\\[2pt]
Pick no more than \code{N} distinct specialties best suited for this case. Never add unnecessary specialties just to complete the size.\\[4pt]
Return \textbf{only} a valid JSON array. Each item must be:
\begin{lstlisting}[style=wraptt]
{
  "specialty": "<one from the pool>",
  "role": "<short role name, exactly one item has 'leader'>",
  "description": "<2-4 sentences instructing how this specialist
  should reason for THIS case, including differential focus,
  red flags, and collaboration style>"
}
\end{lstlisting}
No prose outside JSON. No special characters.
\end{promptbox}

\begin{promptbox}{Specialist Opinion Prompt}
\small\setstretch{1.2}\RaggedRight
You are \code{\{role\}} in an MDT.\\
\textbf{Goal:} \code{\{goal\}}\\
\textbf{Patient info:} \code{\{question\}}\\[4pt]
\textbf{HARD CONSTRAINTS:}
\begin{itemize}
  \item Base reasoning strictly on Goal and Patient info.
  \item DO NOT introduce or infer external facts, tests, or treatments.
  \item DO NOT copy reasoning from other roles.
  \item If information is missing, write ``insufficient evidence'' instead of guessing.
\end{itemize}
\textbf{OUTPUT FORMAT (STRICT):}
\begin{lstlisting}[style=wraptt]
1) Reflection (2-3 sentences)
2) <diagnosis> block with exactly 10 numbered items:
   1. [Diagnosis 1]: [1-2 sentence rationale]
   ...
   10. [Diagnosis 10]: [1-2 sentence rationale]
</diagnosis>
\end{lstlisting}
\end{promptbox}

\begin{promptbox}{Peer Review Prompt}
\small\setstretch{1.2}\RaggedRight
You are a \code{\{reviewer role\}} in a multidisciplinary team (MDT). Your task is to provide a critical peer review of the \code{\{target role\}}’s opinion on the current case.\\[4pt]
\textbf{Output strictly as JSON:}
\begin{lstlisting}[style=wraptt]
{
  "analysis": "3-4 sentences, concise expert analysis
               highlighting agreements and critical disagreements.",
  "agreements": [
    "Specific, constructive point of concurrence.",
    "Another key area of agreement with rationale."
  ],
  "disagreements": [
    "Precise point of contention with reasoning.",
    "Another targeted critique."
  ]
}
\end{lstlisting}
If you fully concur, set "disagreements" to \code{null}.
\end{promptbox}

\begin{promptbox}{Final Decision Prompt}
\small\setstretch{1.2}\RaggedRight
You are the chair of an MDT. Read the case snapshot and reason step by step.\\[2pt]
\textbf{MDT Investigations Summary:} \code{\{assessment report\}}\\
\textbf{Question:} \code{\{question\}}\\[4pt]
\textbf{Output format:}
\begin{lstlisting}[style=wraptt]
<analysis>
- Summarize key findings and red flags.
- Explain differential logic and tie-breakers.
</analysis>

<top10>
[1] <Diagnosis name>
[2] <Diagnosis name>
...
[10] <Diagnosis name>
</top10>
\end{lstlisting}
Provide exactly 10 diagnoses. No extra text outside \code{<analysis>} and \code{<top10>}.
\end{promptbox}

\subsection{Rubrics for LLM Judge in Agent's Utterance}
\label{rubrics}

This rubric converts free-form predictions into a single categorical judgment for evaluation. The instructions prefer clinical synonymy while rejecting incompatible subtypes, balancing sensitivity and specificity for leaderboard scoring.

\begin{promptbox}{Evaluation Prompt}
\small\setstretch{1.25}\RaggedRight
I will now give you ten predicted diseases. If the predicted diagnosis is in the standard diagnosis list, output its rank (1–10); otherwise, output \code{"No"}. Output exactly one value—either \code{"No"} or a single number from 1 to 10. If multiple match, choose the highest rank.\\[4pt]
Decide whether the predicted disease and a standard diagnosis are the \textbf{same medical condition}. Be moderately strict: allow synonyms and parent unspecified subtype matches, but do not accept clearly distinct subtypes or genetic forms as the same.\\[6pt]
\textbf{Matching Rules:}
\begin{itemize}
  \item \textbf{ACCEPT} if they are synonyms, eponyms, abbreviations, or different wording for the same condition.
  \item \textbf{ACCEPT} if one is a broad parent category and the other is an unspecified form within that category.
  \item \textbf{REJECT} if they specify different subtypes (e.g., type I vs type II), different enzyme defects, or different genes.
  \item \textbf{REJECT} if they are unrelated conditions or only partially overlapping.
\end{itemize}
\textbf{Output Format:}
\begin{lstlisting}[style=wraptt]
<think>
Step-by-step reasoning:
</think>
<answer>No|1-10</answer>
\end{lstlisting}
\textbf{Example Input:}\\
Predicted diseases: \code{\{predict\_diagnosis\}}\\
Standard diagnosis: \code{\{golden\_diagnosis\}}
\end{promptbox}

\subsection{Prompts for LLM Summarizer}
\label{summarizer}
The summarizer condenses multi-turn MDT content into an actionable brief for clinicians or downstream modules, emphasizing signal over verbosity and avoiding speculative language.

\begin{promptbox}{Prompts for MDT Chair Summarizer}
\small\setstretch{1.2}\RaggedRight
You are the chair of an MDT. Read the case snapshot and reason step by step.

MDT Investigations Summary:
\code{\{assessment\_report\}}

Question: \code{\{question\}}

Output format (STRICT):
\texttt{<analysis>}
- Summarize key findings and red flags.
- Differential logic: why certain classes of diseases are prioritized.
- Tie-breakers and evidence weighting.
\texttt{</analysis>}

\texttt{<top10>}
[1] \textless Diagnosis name\textgreater
[2] \textless Diagnosis name\textgreater
[3] \textless Diagnosis name\textgreater
[4] \textless Diagnosis name\textgreater
[5] \textless Diagnosis name\textgreater
[6] \textless Diagnosis name\textgreater
[7] \textless Diagnosis name\textgreater
[8] \textless Diagnosis name\textgreater
[9] \textless Diagnosis name\textgreater
[10] \textless Diagnosis name\textgreater
\texttt{</top10>}

Rules:\\
- Provide exactly 10 diagnoses, one per line, each starting with [rank].\\
- Be precise and avoid variants on the same concept unless clinically distinct.\\
- No extra text outside \texttt{<analysis>} and \texttt{<top10>}.
\end{promptbox}

\subsection{Retrieval Implementation Details} \label{sec:retrieval_impl}
We implement the retrieval module $\mathcal{M}$ using a dense vector index to inject relevant reasoning priors. Specifically, we employ \texttt{Qwen/Qwen3-Embedding-4B} as the backbone encoder $E(\cdot)$. To ensure the inner product search is equivalent to cosine similarity, we apply $L_2$ normalization to the embeddings of all key-value experience pairs $(k_i, v_i)$ stored in the database, yielding index vectors $\mathbf{u}_i = E(k_i)/\|E(k_i)\|_2$, which are stored using the \texttt{FAISS} library's \texttt{IndexFlatIP}. During inference at time $t$, the current agent's instruction $x_t$ is encoded into a normalized query vector $\mathbf{q}_t = E(x_t)/\|E(x_t)\|_2$. The system retrieves the top-$K$ entries (default $K=8$) by maximizing the similarity score $s_i = \mathbf{q}_t^\top \mathbf{u}_i$ and appends them to the prompt using a strict ``\texttt{EXPERIENCE HINTS}'' template to guide the model's reasoning.

\section{Mathematics}
\label{APP-Mathematics}
\subsection{Detailed Setup}
\label{Math-settup}
\begin{figure*}
    \centering
    \includegraphics[width=1\linewidth]{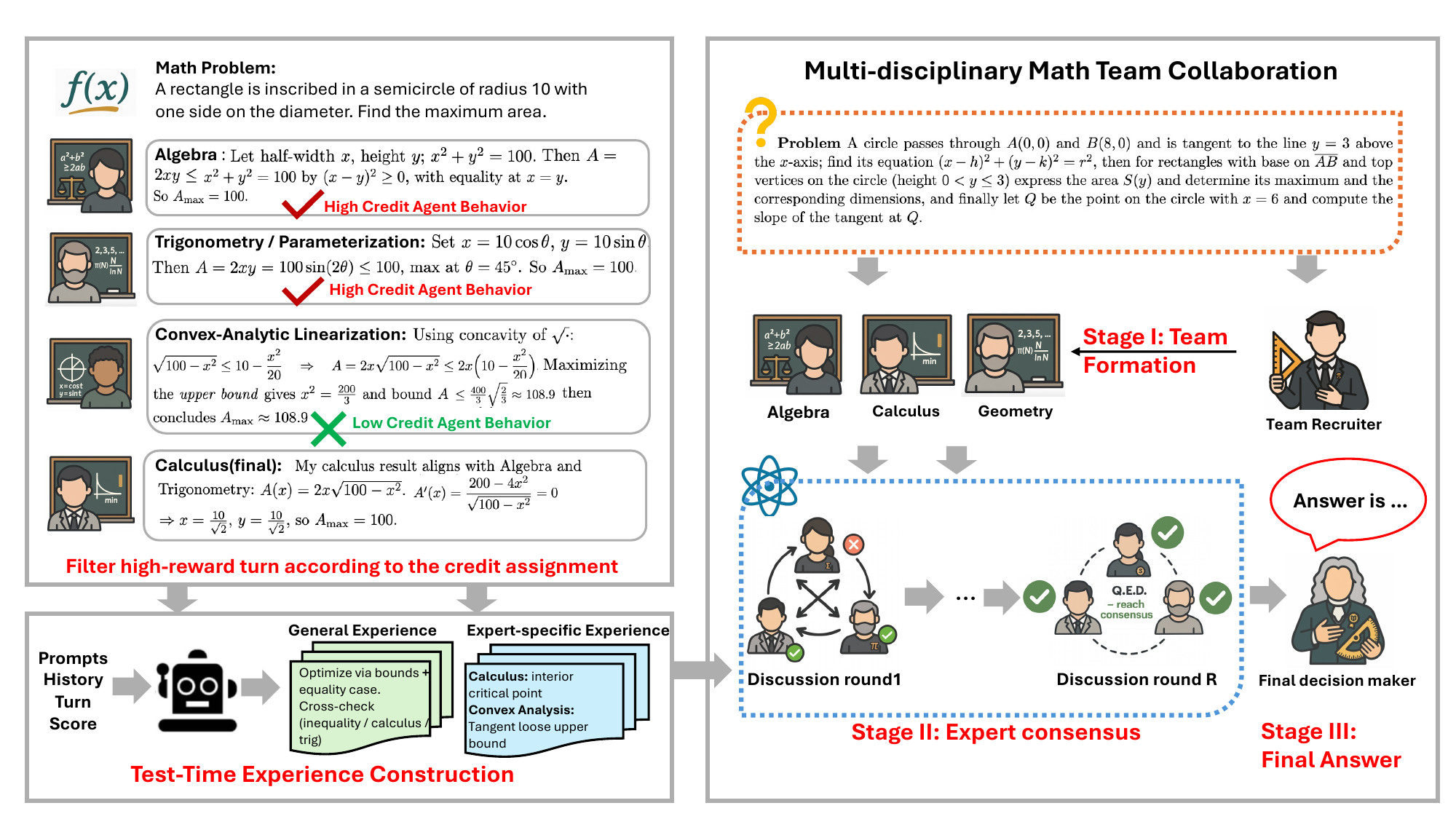}
    \caption{MATTRL in Math: Multi-Specialist Math Problem-solving Collaboration. }
    \label{fig:mattrl_math}
\end{figure*}

We instantiate MATTRL for multi-agent mathematical problem solving (Figure~\ref{fig:mattrl_math}). 
Given a math problem (task record $\mathcal{X}$), the coordinator agent $\mathrm{LLM}_{\mathrm{Coo}}$ forms a small team of specialists, runs a bounded multi-round collaboration with optional experience retrieval, and finally synthesizes a discussion report and outputs the final solution. This appendix specifies the concrete collaboration protocol and prompts used in the math setting.

\paragraph{Baseline run (no experience)}
We first run MATTRL (math) with experience augmentation disabled (\texttt{--use\_experience} off). For each problem, the pipeline outputs a final solution artifact and a detailed interaction log recording each specialist opinion, peer review, and round summary.

\subsection{Free recruitment team formation}
In math, because of the flexibility of math problem, we use free recruitment: instead of selecting from a fixed catalog, the coordinator directly proposes a small set of specialist descriptions tailored to the current problem, and forms $\mathrm{TEAM}$ accordingly. This corresponds to the team-formation stage of our pipeline, where the coordinator constructs a small set of role-specialized agents on-the-fly for each problem.

\begin{promptbox}{Recruitment Prompt for Free Recruitment}\label{prompt:math-recruit-free}
\small\setstretch{1.2}\RaggedRight
You are the Chief Problem Solving Officer assembling a small math team.\\[2pt]
Problem: \code{\{problem\}}\\[4pt]
Design up to \code{\{n\}} distinct specialties best suited for this problem.\\
You are NOT restricted to any preset pool. Create precise specialist titles that reflect reasoning roles (e.g., invariant designer, structure normalizer, edge-case auditor).\\
Exactly ONE item must have role \code{leader}.\\[6pt]

Return ONLY a valid JSON array. Each item must be:
\begin{lstlisting}[style=wraptt]
{
  "specialty": "<your created specialist title>",
  "role": "<short role name, exactly one item has 'leader'>",
  "description": "<2-4 sentences on how this specialist will reason for THIS problem:
                  key invariants, typical tactics, how to communicate>"
}
\end{lstlisting}
No prose outside JSON.
\end{promptbox}

\subsection{Multi-round collaboration (Stage II)}
We run up to $R_{\max}$ collaboration rounds. In each round, every non-converged specialist proposes a solution attempt; other specialists then provide targeted critiques and minimal fixes in a structured format. The coordinator aggregates these critiques into a concise feedback bulletin, which is provided to specialists in the next round for revision. A specialist is marked converged once their solution no longer changes under critique.

\begin{promptbox}{Round 1 Specialist Opinion Prompt}\label{prompt:math-opinion-r1}
\small\setstretch{1.25}\RaggedRight
You are \code{\{specialty\} (\{role\})} in a math problem-solving team.\\
Goal: \code{\{team\_goal\}}\\
Problem: \code{\{problem\}}\\[6pt]
Please think step by step from your expert perspective, and produce ONE integrated, concise solution message.\\
Include: brief reflection (1--2 sentences), core reasoning leading to the result, and the final answer (one line if known).\\
Keep it rigorous and self-contained; avoid decorative tags.
\end{promptbox}

\subsection{Structured peer review and acceptance rule}
For each specialist’s attempt, all other specialists generate a structured peer review in raw JSON, including an overall appraisal, a verdict (\texttt{accept}/\texttt{revise}/\texttt{reject}), validated parts, and a list of concrete issues with severities (\texttt{fatal}/\texttt{major}/\texttt{minor}) and minimal fixes. A specialist’s attempt is marked accepted \emph{only if} (i) all peer verdicts are \texttt{accept} and (ii) the \texttt{issues} list is empty. When critiques identify no remaining issues, we treat the specialist’s update as converged (i.e., no further changes are proposed in subsequent rounds). The collaboration halts when all specialists converge or when reaching round budget.

\begin{promptbox}{Structured Peer Review Prompt }\label{prompt:math-peer-review}
\small\setstretch{1.25}\RaggedRight
You are reviewing another specialist's solution attempt from your own perspective.\\[4pt]
Context bundle (JSON):\\
\code{\{bundle\_json\}}\\[6pt]
DISCIPLINE:\\
- Keep \code{analysis} to 3--4 concise sentences: start with grounded agreements, then targeted critiques.\\
- Put parts you believe correct in \code{validated} as short phrases.\\
- For EACH critique, add an entry in \code{issues} with type, severity, concrete note, and a minimal fix.\\
- Use \code{tests} to note quick sanity checks (toy examples, boundary cases, counterexamples).\\[6pt]
Verdict rule: \code{accept} only if there are NO issues at any severity (including minor) and the solution is essentially complete;\\
\code{revise} if ANY fixable issue remains; \code{reject} if there is a fatal flaw.\\[6pt]
OUTPUT (RAW JSON ONLY):
\begin{lstlisting}[style=wraptt]
{
  "analysis": "2-4 sentences overall appraisal.",
  "verdict": "accept | revise | reject",
  "validated": ["..."],
  "issues": [
    {"type":"algebra_error|logic_gap|missing_case|theorem_misuse|undefined_symbol|diagram_dependence...",
     "severity":"fatal|major|minor",
     "note":"Exactly what is wrong.",
     "fix":"Concrete minimal correction consistent with the current approach."}
  ],
  "tests": {"random_checks": 0, "counterexample": {"found": false, "example": null}},
  "alt_hint": "Optional short hint to guide a revision.",
  "confidence": 0
}
\end{lstlisting}
\end{promptbox}

\begin{promptbox}{Revision Opinion Prompt with Peer Feedback}\label{prompt:math-opinion-revise}
\small\setstretch{1.25}\RaggedRight
You are \code{\{specialty\} (\{role\})} in a math team.\\
Problem: \code{\{problem\}}\\[4pt]
Last accepted snapshot (may be empty):\\
\code{\{accepted\_snapshot\}}\\[4pt]
Peer feedback for YOU (JSON):\\
\code{\{delta\_json\}}\\[6pt]
Please think step by step from your expert perspective, and produce ONE integrated, concise message addressing feedback (no step numbering).\\
State the refined reasoning and the final answer if applicable.\\
Be precise and minimal; no special tags.
\end{promptbox}

\subsection{Chair aggregation (final decision)}
After bounded discussion, the coordinator $\mathrm{LLM}_{\mathrm{Coo}}$ synthesizes a discussion report $\mathrm{DR}$ from all specialists’ updates (Stage III), and outputs the final solution. If the first chair output does not contain these tags, the system triggers a rewrite pass that preserves mathematical content but enforces the target format.

\begin{promptbox}{Chair Aggregation Prompt}\label{prompt:math-chair}
\small\setstretch{1.25}\RaggedRight
You are the chair of a math MDT. Integrate accepted outlines and produce a concise, rigorous write-up.\\[4pt]
Snapshot summary (accepted/unresolved):\\
\code{\{assessment\_report\}}\\[6pt]
Problem:\\
\code{\{problem\}}\\[6pt]
OUTPUT FORMAT (STRICT):
\begin{lstlisting}[style=wraptt]
<analysis>
- Key ideas and why this route is effective.
- Note any edge cases handled.
</analysis>

<final_answer>
- Single line with the final numeric/symbolic answer; if unknown, write 'N/A'.
</final_answer>

<formal_proof>
Provide a clean, step-by-step solution/proof.
</formal_proof>
\end{lstlisting}
No extra text outside these tags.
\end{promptbox}

\subsection{Rubrics for LLM Judge in Math Utterance Scoring}
\label{app:math:utterance_scoring}

We use an LLM judge to score (i) the terminal correctness of the final answer and
(ii) the per-utterance contribution within the multi-agent transcript.
The terminal judgment provides the team outcome signal $G\in\{0,1\}$, while the
utterance-level score $s_{i,t}$ measures how much a given agent utterance helps
(or hurts) reaching the correct final solution. In implementation, the judge outputs
an integer score in $[0,5]$, and we optionally normalize it to $[0,1]$ by
$s_{i,t}=\mathrm{score}/5$.

\begin{promptbox}{Math Judge Prompt: Terminal Answer Correctness}
\small\setstretch{1.15}\RaggedRight
You are a strict math judge. Determine whether the predicted final answer matches
the golden answer (allowing standard mathematical equivalence).\\[4pt]
\textbf{Problem:} \code{\{problem\}}\\
\textbf{Predicted Final Answer:} \code{\{final\_answer\}}\\
\textbf{Golden Answer:} \code{\{golden\}}\\[4pt]
Output format (STRICT):
\begin{lstlisting}[style=wraptt]
<analysis>
Briefly justify equivalence/non-equivalence.
</analysis>
<answer>
Yes|No
</answer>
\end{lstlisting}
\end{promptbox}

\begin{promptbox}{Math Judge Prompt: Utterance Influence Scoring (0--5)}
\small\setstretch{1.15}\RaggedRight
You are a math reasoning evaluator. Your task is to score how much the target
agent utterance influences reaching the correct final answer.\\
Score range: 0 (no helpful influence) to 5 (decisive helpful influence).\\[4pt]

Context you will be given:\\
- Problem statement\\
- Step 0 snapshot (baseline)\\
- Step i snapshot (with the target utterance)\\
- Predicted final answer and golden answer\\
- The target agent's prompt and response\\[6pt]

Output STRICT JSON only:
\begin{lstlisting}[style=wraptt]
{
  "analysis": "1-3 sentences explaining why this utterance helps/hurts.",
  "score": 0-5
}
\end{lstlisting}
\end{promptbox}

We then combine the utterance score with the decayed terminal signal:
the terminal correctness $G$ is allocated to turns with a decay factor and
distributed among agents within the same turn proportionally to their utterance
scores, and finally fused with the direct utterance score to obtain $r_{i,t}$
used for experience selection.
\subsection{Interaction scoring and selection (train split only)}
We score each specialist utterance with an LLM judge to obtain an individual score $s_{i,t}$, then combine it with a terminal correctness signal allocated back to turns using a decay factor.
 Each specialist utterance is scored by an LLM judge to obtain an individual score $s_{i,t}$, and a terminal correctness signal $G$ is allocated back to turns with decay. We then select high-value utterances (e.g., top quantile or thresholded by $r_{i,t}$) to form the candidate set for experience extraction.

\subsection{Experience extraction and indexing (train split only)}
Selected high-value utterances are distilled into concise textual experiences using a fixed LLM-based summarization template, producing key--value entries that are easy to retrieve. We embed the keys, build a dense index (Appendix~\ref{sec:retrieval_impl}), and retrieve top-$K$ experiences at inference time. Retrieved experiences are appended to prompts using the standardized \texttt{EXPERIENCE HINTS} block.

\subsection{Test-Time Experience Retrieval}
\label{math-retrieval}
At test time, each non-converged specialist retrieves relevant experiences from the shared pool $\mathcal{E}$ based on the current problem and its round context. Retrieval is implemented with dense embeddings and a FAISS index. The retrieved entries are appended to the prompt using the same \texttt{EXPERIENCE HINTS} template as other domains, serving as consultable guidance without updating model weights.

\begin{promptbox}{Test-time Experience Hint Injection }\label{prompt:math-exp-hints}
\small\setstretch{1.25}\RaggedRight
When experience augmentation is enabled, retrieved hints are appended to the end of the prompt:\\[4pt]
\begin{lstlisting}[style=wraptt]
===== EXPERIENCE HINTS (consult; do not quote verbatim) =====
- ACTION: <retrieved key 1>
  EXPERIENCE: <retrieved experience 1>
- ACTION: <retrieved key 2>
  EXPERIENCE: <retrieved experience 2>
...
===== END OF EXPERIENCE HINTS =====
\end{lstlisting}
Agents may consult these hints to improve reliability (e.g., invariant choice, missing cases, counterexample checks),
but should not quote them verbatim in the final solution.
\end{promptbox}

\section{Education}
\label{APP-Education}

\subsection{Detailed Setup}
\label{education-setup}
\begin{figure*}
    \centering
    \includegraphics[width=1\linewidth]{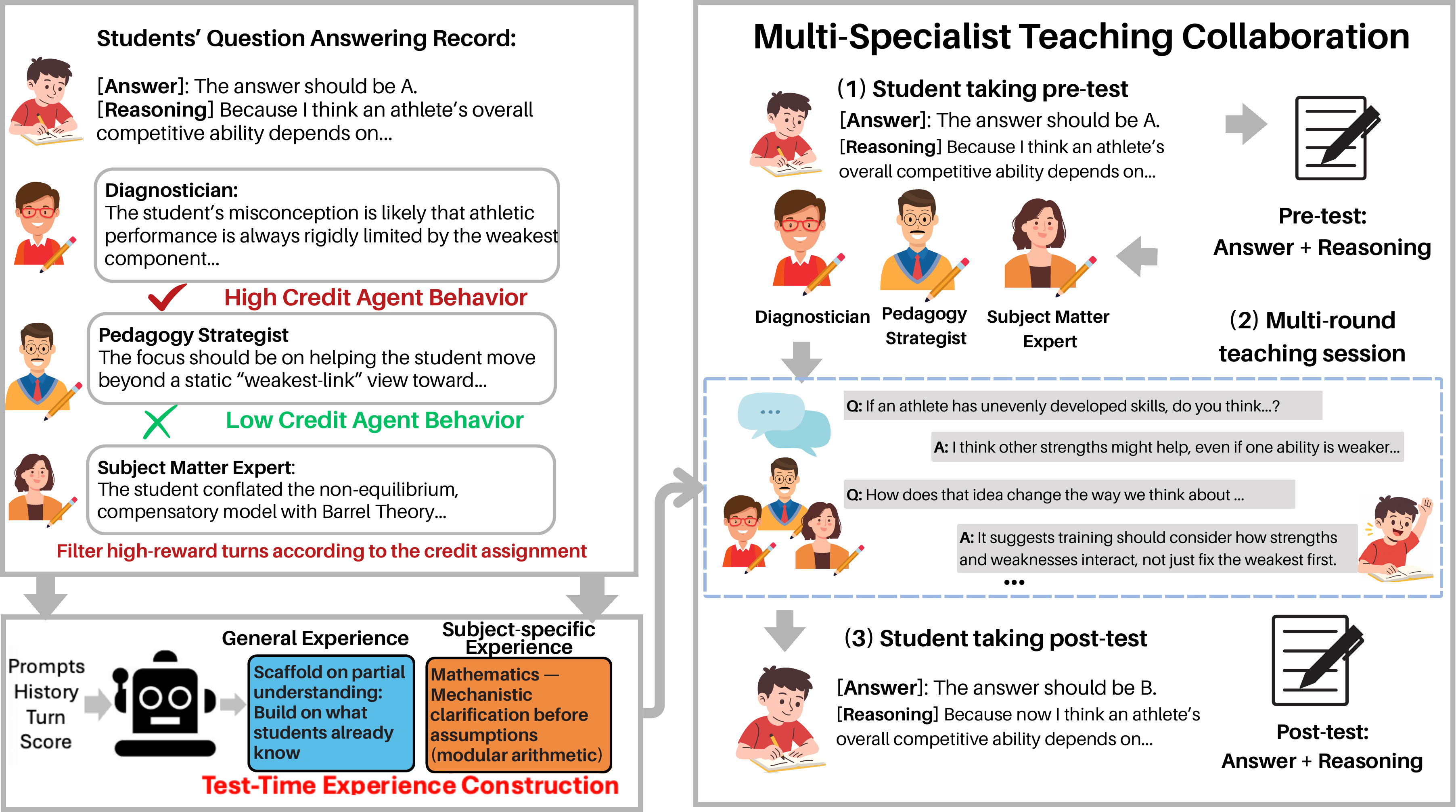}
    \caption{MATTRL in Education: Multi-Specialist Teaching Collaboration. }
    \label{fig:mattrl_edu}
\end{figure*}

Large language models are increasingly serving as educational tools, yet evaluating their teaching capabilities remains challenging. In this experiment, we adapt the MATTRL framework and create a realistic learning scenario in which a team of pedagogy specialists works together to guide students through complex problem-solving tasks (Figure \ref{fig:mattrl_edu}). This setup allows us to test how effective the MARRLL is at improving the teaching performance of multi-agent systems.
\paragraph{Pre-test}
A pre-test is conducted to establish baseline student performance before any instruction. A student agent (GPT-4o, temperature=0.3) is prompted (in \ref{pre-test-prompt}) to solve multiple-choice questions from SuperGPQA, providing both the answer and reasoning to surface its thinking and uncertainties. Pre-test questions are selected via stratified sampling across 13 subject matters and three difficulty levels to ensure balanced coverage. The pre-test is run once before any teaching sessions, and the same student agent instance is reused across all experimental conditions.

\paragraph{Pedagogy Specialist Team Formation}
Before the instructional session, a pedagogy specialist team of three members is formed based on an analysis of the question and the students’ pre-test performance. Team members are selected from a predefined pool (\ref{Pedagogy Specialist Recruitment}, Table~ \ref{tab:specialist-pool}) that includes subject-matter experts, pedagogical specialists, and cross-disciplinary specialists. Each team member is assigned a specific role: the diagnostician identifies the reasons for the student’s incorrect response, the pedagogy strategist proposes appropriate instructional strategies, and the subject matter expert provides discipline-specific explanations.

\paragraph{Multi-round teaching session}\label{teaching_session}
During the teaching session, the teaching agent (GPT-5, temperature=0.3) is provided with the full question text and the correct answer, the student’s pre-test response and reasoning, and the correctness status. The teacher agent guides the student toward the correct answer through a structured, three-round question–answer dialog that diagnoses and clarifies misconceptions while scaffolding the student’s reasoning, without directly revealing the answer. Three teaching conditions are evaluated for comparison: (1) a \textit{Single-Teacher} condition, in which a single agent conducts the full dialog using a fixed instructional prompt (\ref{prompt:single-teacher}); (2) a \textit{Multi-Teacher} condition, in which multiple specialist agents generate each instructional strategy analysis based on their role-specific perspectives first and collaboratively plan before interacting with the student agent ( Prompt:~\ref{prompt:multi-teacher}); and (3) a \textit{Multi-Teacher with Experience} condition, which extends the collaborative setting by incorporating role-, subject-, and difficulty-specific teaching experiences retrieved from the experience pool to inform instructional strategy generation (Prompt:~\ref{prompt:multi-teacher-exp}).

\paragraph{Post-test}
In the post-test, the student agent answers the same question again using the same response format as the pre-test. If the student answered correctly on the pre-test, the teaching session will be skipped, and the pre-test answer will be reused in the post-test.

\paragraph{Interaction scoring and selection}
To construct the pedagogy experience pool, additional teaching interactions are generated using stratified sampling over subject domains and difficulty levels from the SuperGPQA dataset under the multi-agent teaching setting described above. 28 successful cases are finally identified and scored using two complementary signals. First, a global outcome score captures overall instructional success and is defined as a binary indicator of post-test correctness, assigning a value of $1.0$ if the student’s post-test answer is correct and $0.0$ otherwise. Second, a step-level influence score evaluates the contribution of each teaching-strategy utterance to student learning. Each utterance is rated on a $0$--$5$ scale by an LLM adjudicator, measuring its causal influence on the student’s progress relative to the pre-test baseline. In addition, each role of teachers' pedagogy analyzing utterance is evaluated using a rubric-based utterance quality score (\ref{utterance scoring}). The binary global outcome score is temporally allocated across dialogue turns using decay with factor $\gamma = 0.85$, assigning higher credit to earlier instructional turns. Within each turn, the allocated global credit is distributed across utterances in proportion to their step-level influence scores. Finally, each utterance is assigned a combined score computed as a weighted average of its share of the decayed global credit and its direct instructional contribution, with weights of 0.6 and 0.4, respectively.

\paragraph{Experience extraction and summarization}
From each scored teaching interaction, the top-ranked (25\%) utterances are selected based on their \textit{final\_score} and converted into transferable pedagogical experiences using an LLM-based extractor (\ref{LLM Summarizer}). Each extracted experience follows a constrained instructional format and is categorized as either general or subject-specific. Experiences are indexed by the teacher role, subject domain, and difficulty level, and stored in a structured format. We provide example experiences here in ~\ref{experience_examples}.

\subsubsection{Test-Time Experience Retrieval}
\label{edu-retrieval}
At test time, when experience augmentation is enabled, each teacher agent first attempts to load a role-specific pedagogy experience knowledge base according to the corresponding question subject matter and difficulty level. The role-specific knowledge base is identified using the agent’s assigned instructional role (e.g., Diagnostician, Subject Matter Expert, or Pedagogy Strategist). Retrieved experiences are appended to the agent’s prompt in a \emph{Experience Hints} section, explicitly marked as consultative guidance intended to inform the agent’s instructional decisions, rather than to be quoted verbatim in generated responses.

\subsection{Experience Examples}\label{experience_examples}

\begin{figure}[ht]
\centering
\begin{tikzpicture}
    \node[mybox, inner sep=3pt] (box){%
        \begin{minipage}{0.98\columnwidth} 
        \scriptsize 
        \setstretch{1.05} 
        \setlength{\parskip}{0pt}
        \setlength{\parindent}{0pt}

        \begin{flushleft}
        \vspace{2mm}
        \textbf{General Teaching Experience: }
        
        \includegraphics[width=0.28cm]{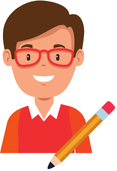}\hspace{4pt}\textbf{Scaffold on partial understanding}: Identify and leverage what students already understand, using existing knowledge as a foundation for introducing new concepts.\\
        \includegraphics[width=0.28cm]{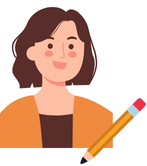}\hspace{4pt}\textbf{Anchor on key discriminators first}: Organize instruction around the core distinctions or decision points that differentiate correct from incorrect reasoning, using them as the backbone of explanation.\\
        \includegraphics[width=0.28cm]{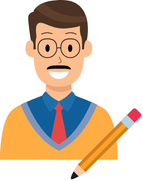}\hspace{4pt}\textbf{Explicitly name misconceptions}: Surface and label specific misconceptions directly, rather than correcting them implicitly, to prevent persistent or repeated errors.\\

        \vspace{2mm}
        \textbf{Subject-Specific Teaching Experience: }

        \includegraphics[width=0.28cm]{assets/teacher1.png}\hspace{4pt}\textbf{Mathematics}: Prioritize mechanistic clarification before introducing assumptions, particularly in structured domains such as modular arithmetic, where procedural misunderstandings often precede conceptual errors.\\
        \includegraphics[width=0.28cm]{assets/teacher2.png}\hspace{4pt}\textbf{Philosophy}: Emphasize precise definitions and logical scope, especially in tasks involving entailment, quantifiers, or closely related concepts, where minor definitional shifts can alter reasoning outcomes.\\
        \includegraphics[width=0.28cm]{assets/teacher3.png}\hspace{4pt}\textbf{Medicine}: Apply a high evidentiary standard when distinguishing descriptive from causal claims, with particular attention to study design and classification to avoid overgeneralization or misinterpretation.\\
        \end{flushleft}
        \end{minipage}
    };

    \node[fancytitle, rounded corners, right=8pt] at (box.north west) {Experiences};
\end{tikzpicture}
\vspace{-6mm}
\caption{General \& subject-specific experience}
\label{experience}
\end{figure}

\subsection{Description of Specialist Pool}\label{tab:specialist-pool}
This specialist pool spans key academic domains, pedagogical expertise, and cross-disciplinary perspectives, enabling flexible and targeted formation of specialist teams for instructional support.

\begin{table}[h!]
\centering
\footnotesize
\setlength{\tabcolsep}{4pt}
\renewcommand{\arraystretch}{1.15}
\begin{tabular}{p{0.13\textwidth} p{0.32\textwidth}}
\toprule
\textbf{Category} & \textbf{Specialist Pool} \\
\midrule
Domain Experts &
Mathematics; Engineering; Physics; Chemistry; Biology; Computer Science; Medicine; Agriculture; Economics; Management; Law; Education; Military Science; History; Literature; Philosophy; Sociology; Language Arts \\
Pedagogical Specialists &
Pedagogy; Educational Psychology; Assessment and Evaluation; Curriculum Design \\
Cross-Disciplinary Specialists &
STEM; Humanities; Social Sciences \\
\bottomrule
\end{tabular}
\caption{Specialist pool used for pedagogy team formation.}
\end{table}

\subsection{Prompts}

\subsubsection{Prompt for Student Agent in Pre-test}\label{pre-test-prompt}
This prompt guides the student agent to answer a multiple-choice question while explicitly articulating its reasoning process.
\begin{promptbox}{Prompt for Student Agent in Pre-test}
\small\setstretch{1.25}\RaggedRight
Answer the following multiple-choice question. There is only one correct answer. \\

Before giving the final answer, briefly explain: (1) Your reasoning for how the correct option is determined (2) Any uncertainties, ambiguities, or alternative interpretations you considered.

The last line of your response should be in the format:

Answer: \$LETTER' (without quotes), 

(where LETTER is one of A, B, C, D, E, F, G, H, I, or J.

[Question text]

A) [Option A]

B) [Option B]

...

\end{promptbox}

\subsubsection{Pedagogy Specialist Recruitment Prompt}\label{Pedagogy Specialist Recruitment}
This prompt guides the pedagogy specialist to assemble an appropriate teaching team by identifying the pedagogical expertise required for the given pre-test question.
\begin{promptbox}{Pedagogy Specialist Recruitment Prompt}\
\small\setstretch{1.2}\RaggedRight
You are an experienced educational coordinator assembling a teaching team.\\[2pt]
\textbf{Pre-test Question:} \code{\{question\}}\\[2pt]

IMPORTANT: Your team MUST include these key pedagogical roles:
1. The Diagnostician: Analyzes why the student failed the pre-test (e.g., calculation error vs. conceptual misunderstanding)
2. The Pedagogy Strategist: Proposes the next move (e.g., 'Ask a probing question' vs. 'Provide a counter-example')
3. Subject Matter Experts: Provide disciplinary knowledge explanation (select 1-2 from the pool based on question topic)\\[2pt]

From this teacher/specialist pool: \code{\{POOL\}}\\[2pt]
Pick no more than {n} distinct specialties best suited for this question. 
Assign roles so that you have:

- At least one Diagnostician role (can assign to any specialty, but they focus on analyzing student errors)

- At least one Pedagogy Strategist role (can assign to any specialty, but they focus on teaching strategy)

- 1-2 Subject Matter Experts (actual domain specialists from the pool)

- Optionally one lead teacher to coordinate

Never add unnecessary specialties just to complete the size.

Return \textbf{only} a valid JSON array. Each item must be:
\begin{lstlisting}[style=wraptt]
{
  "specialty": "<one from the pool, or use a pool specialty name>",
  "role": "<one of: Diagnostician, Pedagogy Strategist, Subject Matter Expert, lead teacher>",
  "description": "<2-4 sentences instructing an LLM acting as this teacher on how to approach THIS question. 
                  If role is Diagnostician, focus on error analysis. If role is Pedagogy Strategist, focus on teaching methods. 
                  If role is Subject Matter Expert, focus on domain knowledge. Include how to collaborate with other teachers>"
}
\end{lstlisting}
No prose outside JSON. No special characters.
\end{promptbox}

\subsubsection{Prompts for Single-Teacher Instruction}\label{prompt:single-teacher}
This prompt guides the teacher agent to generate instructional feedback based on the student’s pre-test answers and reasoning.
\begin{promptbox}{Prompt for Single-Teacher Instruction}\label{prompt:single-teacher}
\small\setstretch{1.25}\RaggedRight
\textbf{Initialization:}\\
Question: \code{\{question}\}\\
Student's Pre-Test Answer: \code{\{answer}\}\\
Student's Reasoning: \code{\{reasoning}\}\\
Correct Answer: \code{\{gold answer}\}\\
Your task: Guide the student to arrive at the correct answer through a \code{\{num rounds}\}-round dialogue.\\
- You know the correct answer is \code{\{gold answer}\}, but DO NOT directly reveal it\\
- Ask thoughtful questions to guide their thinking toward the correct answer\\
- Address any misconceptions in their reasoning\\
- Help them discover the correct answer through guided discovery\\
- Focus on teaching concepts and reasoning that lead to the correct answer\\
\textbf{Round 1 Prompt:}\\
You are teaching a student who just took a pre-test. Here's what they did:\\
Question:\code{\{question}\}\\
Student's Answer: \code{\{answer}\}\\
Student's Reasoning: \code{\{reasoning}\}\\
Based on the student's answer and reasoning, ask them a thoughtful question to guide their thinking. Ask your question now (just the question, no preamble):\\
\textbf{Student Response Prompt:}\\
The teacher just asked you: \code{\{question}\} Please respond to the teacher's question thoughtfully.\\
\textbf{Round 2 Prompt:}\\
The student just responded to your previous question. Based on their response, ask a follow-up question to continue guiding them. Ask your follow-up question now (just the question, no preamble):\\
\textbf{Final Guidance Prompt:}\\
Now that you've had a dialogue with the student, provide final teaching guidance:\\
- Summarize key concepts they should understand\\
- Clarify any remaining misconceptions\\
- Guide them on how to approach this type of problem correctly\\
- Explain the underlying principles and reasoning process\\
- DO NOT directly state which option/letter is the correct answer\\
\end{promptbox}

\subsubsection{Prompt for Multi-Teacher Instruction}\label{prompt:multi-teacher}
This prompt guides the teacher agent to generate instructional feedback based on the student’s pre-test answer and reasoning.
\begin{promptbox}{Prompt for Multi-Teacher Instruction}
\small\setstretch{1.25}\RaggedRight
\textbf{Round 1 - Individual Analysis Prompt}\\
You are \code{\{role}\}, a specialized teacher analyzing a student's pre-test response.\\
FULL QUESTION:\code{\{question}\}\\
STUDENT'S PRE-TEST RESPONSE:\\
- Answer: \code{\{answer}\}\\
- Reasoning: \code{\{reasoning}\}\\
- Correct: \code{\{correct}\}\\

CORRECT ANSWER: \code{\{gold answer}\} (You know this, but DO NOT reveal it directly to the student)\\

YOUR TASK (based on your role):\code{\{role instructions}\} Analyze the teaching strategy from your perspective. Provide your analysis in 2-3 sentences. 

ROLE-SPECIFIC INSTRUCTIONS:\\
Diagnostician: focus on identifying WHY the student failed\\
Pedagogy Strategist: focus on HOW to teach this student\\
Subject Matter Expert: focus on WHAT domain knowledge the student is missing or misunderstanding\\

\textbf{Round 1 - Collaborative Planning Prompt:}\\
You are a team of specialized teachers collaborating to plan a 2-round teaching dialogue.\\

FULL QUESTION:\code{\{question}\}\\

STUDENT'S PRE-TEST RESPONSE:\\
- Answer: \code{\{answer}\}\\
- Reasoning: \code{\{reasoning}\}\\

CORRECT ANSWER: \code{\{gold answer} (You know this, but DO NOT reveal it directly)\\

INDIVIDUAL TEACHER ANALYSES:\code{\{analyses summary}\}\\

YOUR COLLABORATIVE TASK:\\
Based on all the analyses above, work together to plan specific and targeted questions that will guide the student to discover the correct answer.\\

Requirements:\\
1. Each question should be specific and directly related to the student's reasoning\\
2. DO NOT directly state which option/letter is correct\\

Return your planned questions in this EXACT format: ROUND 1: [specific question here]\\
\textbf{Student Response Prompt:}\\
The teacher just asked you: \code{\{question}\} Please respond to the teacher's question thoughtfully.\\

\textbf{Round 2 - Individual Analysis Prompt}\\

The student just responded to Round \code{\{num}\}:\\

Teacher's Question: \code{\{question}\}\\
Student's Response: \code{\{response}\}\\

Based on your role (\code{\{teacher.role}\}), analyze the teaching strategy from your perspective. Provide your analysis in 2-3 sentences. \\

\textbf{Round 2 - Collaborative Planning Prompt:}\\
You are a team of specialized teachers collaborating to generate the next question for Round \code{\{num + 1}.\\

CONVERSATION HISTORY SO FAR: \code{\{history}\}\\

STUDENT'S LATEST RESPONSE (Round \code{\{num}\}):\code{\{student response}\}\\

INDIVIDUAL TEACHER ANALYSES:\code{\{analyses summary}\}\\

YOUR COLLABORATIVE TASK:\\
Based on all the analyses above, work together to plan a specific and targeted question for Round \code{\{num + 1}\} that will guide the student to discover the correct answer

Return your planned questions in this EXACT format: ROUND 2: [specific question here]\\

\textbf{Final Guidance Prompt:}\\
Now that you've had a dialogue with the student, provide final teaching guidance:\\
- Summarize key concepts they should understand\\
- Clarify any remaining misconceptions\\
- Guide them on how to approach this type of problem correctly\\
- Explain the underlying principles and reasoning process\\
- DO NOT directly state which option/letter is the correct answer\\
\end{promptbox}

\subsubsection{Prompt for Experience Integration}\label{prompt:multi-teacher-exp}
This prompt guides the teacher agent to incorporate retrieved teaching experiences into instruction.
\begin{promptbox}{Prompt for Experience Integration}\label{prompt:multi-teacher-exp}
\small\setstretch{1.25}\RaggedRight
When experience augmentation is enabled, retrieved
hints are appended to the end of the prompt:
\begin{lstlisting}[style=wraptt]
===== USING TEACHING EXPERIENCES =====
Below are teaching experiences from similar successful cases. Use them to INFORM your strategic thinking, but ADAPT them to THIS specific student's error.

HOW TO USE EXPERIENCES:
1. Look for patterns in WHY students made similar errors - does this student's error match?
2. Identify WHERE similar problems occurred - does this help you locate the current student's error?
3. Learn HOW successful teachers addressed similar misconceptions - what strategies worked?
4. Adapt the strategies to THIS student's specific reasoning and error pattern
5. DO NOT blindly copy - each student's error is unique, even if similar

===== EXPERIENCE HINTS (for the role to consult; do not quote verbatim) =====
{experience_hints}

===== END OF EXPERIENCE HINTS =====

REMEMBER: Experiences are GUIDANCE, not rules. Your primary focus is THIS student's specific error, reasoning, and path to the correct answer ({gold_answer}).
\end{lstlisting}

\end{promptbox}

\subsubsection{Prompt for Teaching Utterance Evaluation}\label{utterance_scoring}
This prompt guides an expert evaluator agent to assess a teaching utterance across multiple instructional quality dimensions, including correctness, information gain, relevance, and clarity.
\begin{promptbox}{Prompt for Teaching Utterance Evaluation}\label{utterance scoring}
\small\setstretch{1.25}\RaggedRight
You are an expert educational evaluator.

Evaluate a teacher's utterance based on these criteria:\\

1. CORRECTNESS (30\%): Is the utterance factually correct and accurate?\\
2. INFORMATION GAIN (25\%): Does it provide useful information for student to learn and get closer to the correct answer?\\
3. RELEVANCE (25\%): Is it relevant to the student's misconception or the learning task?\\
4. CLARITY (20\%): Is it clear, understandable, and well-structured?\\

Question: \code{\{question}\}\\
Student's Pre-Test Answer: \code{\{pretest answer}\}\\
Student's Pre-Test Reasoning: \code{\{pretest reasoning}\}\\
Correct Answer: \code{\{gold answer}\}\\

Teacher Utterance:\code{\{teacher utterance}\}\\

Output STRICT JSON:
\begin{lstlisting}[style=wraptt]
{
  "correctness_score": 0-5,
  "information_gain_score": 0-5,
  "relevance_score": 0-5,
  "clarity_score": 0-5,
  "analysis": "<brief explanation of scores>",
  "global_score": 0-5
}

Calculate global_score as weighted sum:
global_score = 0.30 * correctness_score + 0.25 * information_gain_score + 0.25 * relevance_score + 0.20 * clarity_score
Round to nearest integer (0-5).
\end{lstlisting}
When experience augmentation is enabled, retrieved hints are appended to the end of the prompt:
\end{promptbox}

\subsubsection{Prompts for Experience Summarizer}\label{LLM Summarizer}
This prompt guides an experience summarizer agent to extract and structure reusable teaching guidance and strategies from teaching interactions.
\begin{promptbox}{Prompts for Experience Summarizer}\label{LLM Summarizer}
\small\setstretch{1.25}\RaggedRight
You are an expert educational experience analyzer.

TASK: From a {{ROLE, PROMPT, RESPONSE, GROUND TRUTH, EVAL ANALYSIS, FINAL SCORE}} pair, produce retrieval-ready key:value experiences in TWO classes:\\
1) GUIDANCE — how a teacher handles a teaching situation or student error.\\
2) STRATEGY — how a teacher implements a specific teaching approach.\\

INPUTS\\
- PROMPT ROLE: prompt role\\
- INPUT PROMPT: source text containing instructions, constraints, student's answer/reasoning, question context.\\
- INPUT RESPONSE: produced teaching utterance/question/guidance.\\
- GROUND TRUTH (optional): Correct answer\\
- EVAL ANALYSIS (optional): Evaluation analysis of the utterance's effectiveness\\
- FINAL SCORE (optional numeric): Final score of the utterance\\

OUTPUT\\
Return ONE JSON object ONLY where each entry is: "KEY": "EXPERIENCE".\\

KEY CONSTRUCTION (single line, natural English, no brackets):\\
- Format: "{CLASS} — {Role} {action type} {teaching situation} in {Subject} {Difficulty}: {concise content}"\\
- CLASS is either "GUIDANCE" or "STRATEGY"\\
- Role is the teacher's role (e.g., Diagnostician, Pedagogy Strategist, Subject Matter Expert)\\
- teaching situation describes the teaching context (e.g., "misconception", "error analysis", "question generation")\\
- Subject and Difficulty come from the question metadata\\

EXPERIENCE (value) — 1-2 sentences, must:\\
- Be generalizable, instructional teaching guidance (how to identify errors, how to guide students, how to address misconceptions, how to structure effective questions).\\
- Include a clear judgment prefix:\\
  * "Good practice: ..." when behavior aligns with successful teaching outcomes (student corrected their answer, showed understanding).\\
  * "Pitfall: ..." when behavior conflicts with teaching effectiveness or adds confusion.\\
- Optionally append a simple outcome tag at the end: [helpful] / [harmful] / [neutral] / [insufficient].\\
- Justify judgment ONLY by contrasts observable between INPUT RESPONSE and GROUND TRUTH / EVAL ANALYSIS / FINAL SCORE. If these are absent or inconclusive, use "Pitfall/Good practice" only if you can justify from INPUT RESPONSE behavior; otherwise end with [insufficient].\\

\end{promptbox}

\end{document}